\documentclass[10pt,twocolumn,letterpaper]{article}

\usepackage{iccv}
\usepackage{times}
\usepackage{epsfig}
\usepackage{graphicx}
\usepackage{amsmath}
\usepackage{amssymb}
\usepackage{algorithm2e}

\usepackage[pagebackref=true,breaklinks=true,letterpaper=true,colorlinks,bookmarks=false]{hyperref}

\iccvfinalcopy 

\def\iccvPaperID{3065} 

\ificcvfinal\fi
\begin{document}

\title{Causes and Corrections for Bimodal Multipath Scanning with Structured Light}

\author{Yu Zhang, Daniel L. Lau, and Ying Yu\\
University of Kentucky\\
}

\maketitle

\begin{abstract}
\noindent Structured light illumination is an active 3-D scanning technique based on projecting/capturing a set of striped patterns and measuring the warping of the patterns as they reflect off a target object's surface. As designed, each pixel in the camera sees exactly one pixel from the projector; however, there are exceptions to this when the scanned surface has a complicated geometry with step edges and other discontinuities in depth or where the target surface has specularities that reflect light away from the camera. These situations are generally referred to multipath where a given camera pixel receives light from multiple positions from the projector.  In the case of bimodal multipath, the camera pixel receives light from exactly two positions from the projector which occurs when light bounce back from a reflective surface or along a step edge where the edge slices through a pixel so that the pixel sees both a foreground and background surface.  In this paper, we present a general mathematical model and address the bimodal multipath issue in a phase measuring profilometry scanner to measure the constructive and destructive interference between the two light paths, and by taking advantage of this interesting cue, separate the paths and make two separated depth measurements. We also validate our algorithm with both simulation and a number of challenging real cases.    
\end{abstract}

\section{Introduction}
Structured light illumination (SLI) refers to a method of 3D scanning that uses a projector to project a series of light striped patterns such that a camera can reconstruct depth based on the warping of the pattern over the target object's surface~\cite{morano1998structured,geng2011structured,gupta2013structured,gupta2012micro,liu2010dual,gupta2011structured,rosman2016information,o20143d,wang2016dual}.  Examples of SLI include single pattern techniques which project a static pattern that is continuously projected and from which a 3D reconstruction can be made from a single snap shot~\cite{geng2011structured,boyer1987color,geng1996rainbow, freedman2012depth}.  This is the basic approach of the Prime Sense camera used in the Microsoft Kinect V1 product which uses a pseudo-random dot pattern such that dots within small windows of the captured image can be matched to the projected dot constellations~\cite{ryan2016hyperdepth,freedman2012depth,martinez2013kinect}.  

Multiple pattern SLI scanners, alternatively, project a series of patterns, trading temporal resolution for spatial resolution such that each pixel can be independently processed from its neighbors and produce a single point for each pixel in the camera.  As a prime example, the multiple pattern process of gray coding projects the binary bits forming an 8-10 bit address for each pixel of the projector. By deciphering the on-off-on patterns back to binary bits, each pixel can determine what row of the projector they are looking. In Phase Measuring Profilometry, the row coordinates of each pixel are encoded through phase modulation~\cite{srinivasan1984automated,liu2010dual,geng2011structured,chen2008modulated}. These PMP scanners are common for industrial metrology applications with resolutions that can exceed 10 microns.

As an active imaging technique, structured light is susceptible to errors and distortions caused by the redirection of the projected light to form multiple paths from projector to camera besides the direct path of projector to target to camera~\cite{6136522}.  It is a common problem and one of great interest to researchers because of the potentially catastrophic effects on scans. The same problem can be found in a range of 3D imaging modalities like time-of-flight (TOF) where light will reflect off specular surfaces onto neighboring surface points before reflecting back to the camera. Comparing SLI and TOF range modalities, a literature review on the topic of multi-path cancellation reveals an inordinate number of TOF papers over SLI.

Examples of how to deal with multi-path issues in TOF include Dorrington~{\it et al.}~\cite{dorrington2011separating} as well as Bhandari~{\it et al.}~\cite{bhandari2013multifrequency} and Godbaz~{\it et al.}~\cite{godbaz2012closed} who take the common approach of making multiple depth measurements over many different modulation frequencies such that they derive a set of equations from which to fit the phase and magnitude of a multitude of possible component paths. Freedman~{\it et al.}~\cite{freedman2014sra} assume sparsity in reflection and assume the problem is restricted to a small number of multi-path components, which restrict further extension to other scenarios. 

Naik ~{\it et al.}~\cite{naik2015light} take the approach of deriving a light transport model~\cite{Nayar:2006:FSD:1141911.1141977} to combine the standard measurements from a TOF camera with information from direct and global light transport.  By doing so, they separate the phase associated with the direct light path, placing all sub-sequent paths into a single indirect light component.  O'Toole~{\it et al.}~\cite{o20143d,o2014temporal} employ the epipolar geometry constraint and re-design the optical system to separate the direct and indirect light paths. They modify the optical system and block the global component during data capture procedure.  Gupta~{\it et al.}~\cite{gupta2015phasor} study the temporal illumination on and report that global light transport vanishes at high frequencies. They propose a ToF based shape recovery technique and a method to separate direct and global light. Kadambi~{\it et al.}~\cite{kadambi2013coded} use a coded illumination ToF camera to achieve light sweep imaging with multi-path correction. 

Dedrick~\cite{dedrick2011improving} identify multipath in SLI scans without presenting an effective algorithm for extracting the absolute paths from the collected scans. Courture~{\it et al.} ~\cite{couture2011unstructured} design special pattern to overcome interreflections which is quite different from traditional phase shifting pattern. Nayar et al.~\cite{nayar2006fast} show the radiance of a scene point is due to direct illumination of the point by the source and global illumination arising from diffuse interreﬂection, subsurface scattering, volumetric scattering and translucency. Gupta and Nayar~\cite{gupta2012micro} use this conclusion and present an approach using a narrow, high frequency band structured light pattern to separate direct and global illumination for shape recovery for real scenes. However, the separated direct component can still suffer from bimodal multipath. Their method cannot address bimodal multipath in the direct image and will cause severe artifacts in the reconstruction because they still use traditional phase shifting method to solve phase/depth in the direct component. Furthermore, the authors relate that they do not consider the camera defocus effect, resulting in incorrect depths especially at depth edges.

\subsection{Contributions}
In reviewing the available literature on multi-path interference and its cancellation, the many papers devoted to TOF sensing require unique hardware setups which are often times expensive to build. Structure light scanners being readily constructed from commodity components are, therefore, widely studied; however, limited concrete solutions to the multi-path problem exist for these scanners.  In this paper, we present an inexpensive and practical approach to address this issue without any hardware modification by casting the problem of multi-path interference in terms of the constructive and destructive interference of sinusoidal waves of equal frequency commonly associated with the physics of standing waves and moire interferometry. 

The proposed model is consistent with Dorrington~{\it et al.}~\cite{dorrington2011separating}, but it treats the solution in terms of a structured light scanner.  And it includes an intuitive construction that explains how paths interact as a function of the spatial frequency to produce standing waves of constructive and destructive interference.   In so doing, we establish an equation for this interference such that we can visualize multipath as a sinusoidal pattern plotted versus pattern frequency and varying as a function of the phase difference between component paths.  

The experimental results that we present also deal with a problem unique to structured light, and that is the low-pass filtering affect of the component optics that cause high spatial frequency patterns to have a lower amplitude than low spatial frequencies.  In traditional structured light, this is an issue that is largely ignored since the final phase is determined by the high spatial frequency, with lower frequencies used for unwrapping the high.  This paper deals directly with the issue by establishing an envelope function during scanner calibration such that we can observe bi-modal multi-path in the presence of a non-flat spatial frequency response. 

To the best of our knowledge, we are the first to report the interesting constructive and destructive cue for bimodal multipath using signal processing theory and present a practical approach to simultaneously identify and extract the dominant and non-dominant phases/magnitudes by taking advantage of that cue in an intuitive way without any hardware modifications or additional requirements for customized pattern. As a result, it is easy to be integrated with existing structured light systems. Central to this separation, we propose the idea of a zero-frequency PMP pattern which projects a time-varying but spatially constant structured light pattern as a way to observe the modulated light component absent the multi-path interference that my otherwise partially cancel the modulated light. 

\section{Background}
\noindent Three-dimensional surface scanning by means of structured light is performed using a series of striped patterns projected onto a target scene and captured by a digital camera, placed at a triangulation angle of the projector's line of sight.  The pixels of the captured images are then processed to identify a unique projector row coordinate for which the subject camera pixel corresponds. Perhaps one of the simplest means of SLI is through the use of phase-shift keying where the component patterns are defined by the set, $\{I^p_n:n=0,1,\ldots,N-1\}$, according to:
\begin{equation}
    \label{EQ:ProjectorPattern}
    I^p_n(x^p, y^p) = \frac{1}{2} + \frac{1}{2} \cos \left( 2\pi (\frac{n}{N}-y^p)\right).
\end{equation}
where $(x^p, y^p)$ is the column and row coordinate of a pixel in the projector, $I_n^p$ is the intensity of that pixel in a projector with dynamic range from 0 to 1, and $n$ represents the phase-shift index over the $N$ total patterns.

For reconstruction, a camera captures each image where the sine wave pattern is distorted by the scanned surface topology, resulting in the patterned images expressed as:
\begin{equation}
    I_n^c(x^c, y^c) = A^c + B^c \cos\left( \frac{2\pi n}{N} - \theta \right).
    \label{EQ:CameraPattern}
\end{equation}
where $(x^c, y^c)$ is the coordinates of a pixel in the camera while $I_n^c(x^c, y^c)$ is the intensity of that pixel. The term $A^c$ is the averaged pixel intensity across the pattern set that includes the ambient light component, which can be derived according to:
\begin{equation}
    A^c = \frac{1}{N}\sum_{n=0}^{N-1}I_n^c(x^c, y^c).
    \label{EQ:Ac}
\end{equation}
Correspondingly, the term $B^c$ is the intensity modulation of a given pixel and is derived from $I_n^c(x^c, y^c)$ in terms of real and imaginary components where: 
\begin{equation}
    B^c_{\cal R}  =  \sum_{n=0}^{N-1}I_n^c(x^c, y^c) \cos\left(\frac{2\pi n}{N}\right)
    \label{bcCosEqn}
\end{equation}
and
\begin{equation}
    B^c_{\cal I}  =  \sum_{n=0}^{N-1}I_n^c(x^c, y^c) \sin\left(\frac{2\pi n}{N}\right)
    \label{bcSinEqn}
\end{equation}
such that 
\begin{equation}
    B^c  =  \left\| {B^c_{\cal R}}+j{B^c_{\cal I}}   \right\| =  \left\{ {B^c_{\cal R}}^2+{B^c_{\cal I}}^2 \right\}^\frac{1}{2},
\end{equation}
which is the amplitude of the observed sinusoid.

If $I_n^c(x^c, y^c)$ is constant or less affected by the projected sinusoid patterns, $B^c$ will be close to zero. Thus $B^c$ is employed as a shadow noise detector/filter~\cite{Li:97} such that the shadow-noised regions, with small $B^c$ values, are discarded from further processing. Of the reliable pixels with sufficiently large $B^c$, $\theta$ represents the phase value of the captured sinusoid pattern derived as:
\begin{equation}
    \theta = \angle ({B^c_{\cal R}}+j{B^c_{\cal I}}) =   \arctan \left\{\frac{B^c_{\cal I}}{B^c_{\cal R}}\right\},
    \label{EQ:Phase}
\end{equation}
which is used to derive the projector row according to $\theta=2\pi y^p$.

Given that the reconstructed $\theta$ is affected by distortions in the projector/camera such as thermal noise~\cite{Daley:98} or gamma~\cite{Liu:10}, eqn.~(\ref{EQ:CameraPattern}) is commonly modified to include higher spatial frequencies according to:
\begin{equation}
    I^p_n(x^p, y^p) = \frac{1}{2} + \frac{1}{2} \cos \left( 2\pi (\frac{n}{N} - K y^p)\right),
\end{equation}
where $K$ is the number of sinusoidal wavelengths across the projector in any one frame.  These higher frequency scans result in ambiguities in $\theta$ which are resolved by phase unwrapping via lower frequency $K$s.  For instance, one might use three separate scans with $K=1$, $4$, and $16$ using the $K=1$ to unwrap the $K=4$ scan and then using that resulting scan to unwrap the $K=16$ scan.  This procedure results in a scan with ${1/16}^{th}$ the noise of just the $K=1$ scan where $y^p = \theta/(K 2\pi)$.

In choosing $K$, an experienced operator knows that quantization noise in the projector requires that $K$ be selected such that the corresponding wavelength of the spatial sinusoids correspond to integer multiples of $N$ pixels; otherwise, banding artifacts are visible in the reconstruction of $\theta$.  At the same time, larger values of $N$ result in less thermal noise as well as in the elimination of gamma.  So while a small $N$ allows for higher spatial frequency $K$, it also results in high levels of Gaussian noise in $\theta$ while also making $\theta$ susceptible to gamma distortion.  As such, we recommend an $N$ no smaller than 8, meaning a VGA projector is limited to a maximum frequency of $K=60$ with the sinusoid moving 1 pixel with each step in $n$.

\section{Bimodal Multi-Path Model}
\noindent In signal processing, it is often times convenient to assume a sample of an analogue signal is its value at an infinitesimally thin sliver of time, but in fact, a sample is the average value of the signal over a fixed interval in time.  In digital cameras, a pixel collects light over a fixed angle in the horizontal, $\theta$, and vertical, $\phi$.  As such, an accurate model of a pixel is not eqns.~(\ref{bcCosEqn}) and (\ref{bcSinEqn}) but by:
\begin{equation}
    B^c_{\cal R}  =  \sum_{n=0}^{N-1} \int_{\theta}\int_{\phi} I_n^c(\theta, \phi) 
    			    \cos\left(\frac{2\pi n}{N}\right) d\theta d\phi
\end{equation}
and
\begin{equation}
    B^c_{\cal I}  = \sum_{n=0}^{N-1} \int_{\theta}\int_{\phi} I_n^c(\theta, \phi) 
    		          \sin\left(\frac{2\pi n}{N}\right) d\theta d\phi.
\end{equation}
In this form, we can now identify the principal problem of multi-path, which occurs when $I_n^c(\theta, \phi)$ corresponds to a foreground object for same range on $\theta$ and $\phi$ and a background object for the rest of the $\theta$ and $\phi$ within the field of view of the subject pixel. We can describe this mathematically according to:
\begin{eqnarray}
    B^c_{\cal R}  & = & \sum_{n=0}^{N-1} \int_{\theta_f}\int_{\phi_f} I_n^c(\theta, \phi) 
    			    \cos\left(\frac{2\pi n}{N}\right) d\theta_f d\phi_f + 					\nonumber \\
& &			     \sum_{n=0}^{N-1} \int_{\theta_b}\int_{\phi_b} I_n^c(\theta, \phi) 
    			    \cos\left(\frac{2\pi n}{N}\right) d\theta_b d\phi_b
\end{eqnarray}
and
\begin{eqnarray}
    B^c_{\cal I}  & = &  \sum_{n=0}^{N-1} \int_{\theta_f}\int_{\phi_f} I_n^c(\theta, \phi) 
    			    \sin\left(\frac{2\pi n}{N}\right) d\theta_f d\phi_f +     					\nonumber \\
& &			     \sum_{n=0}^{N-1} \int_{\theta_b}\int_{\phi_b} I_n^c(\theta, \phi) 
    			    \sin\left(\frac{2\pi n}{N}\right) d\theta_b d\phi_b
\end{eqnarray}
where $\theta_f$ and $\phi_f$ represent the range of $\theta$ and $\phi$ covering the foreground object while $\theta_b$ and $\phi_b$ cover the background object. We can simplify both these equations by writing:
\begin{equation}
    B^c_{\cal R}  =  B^{c,f}_{\cal R} + B^{c,b}_{\cal R}
\end{equation}
and
\begin{equation}
    B^c_{\cal I}  =  B^{c,f}_{\cal I} + B^{c,b}_{\cal I}
\end{equation}
where we added the superscripts $f$ and $b$ to distinguish between the foreground and background components on $B^c_{\cal R}$ and $B^c_{\cal I}$. 

\begin{figure}[!t]
\centering\includegraphics[width=3.25in]{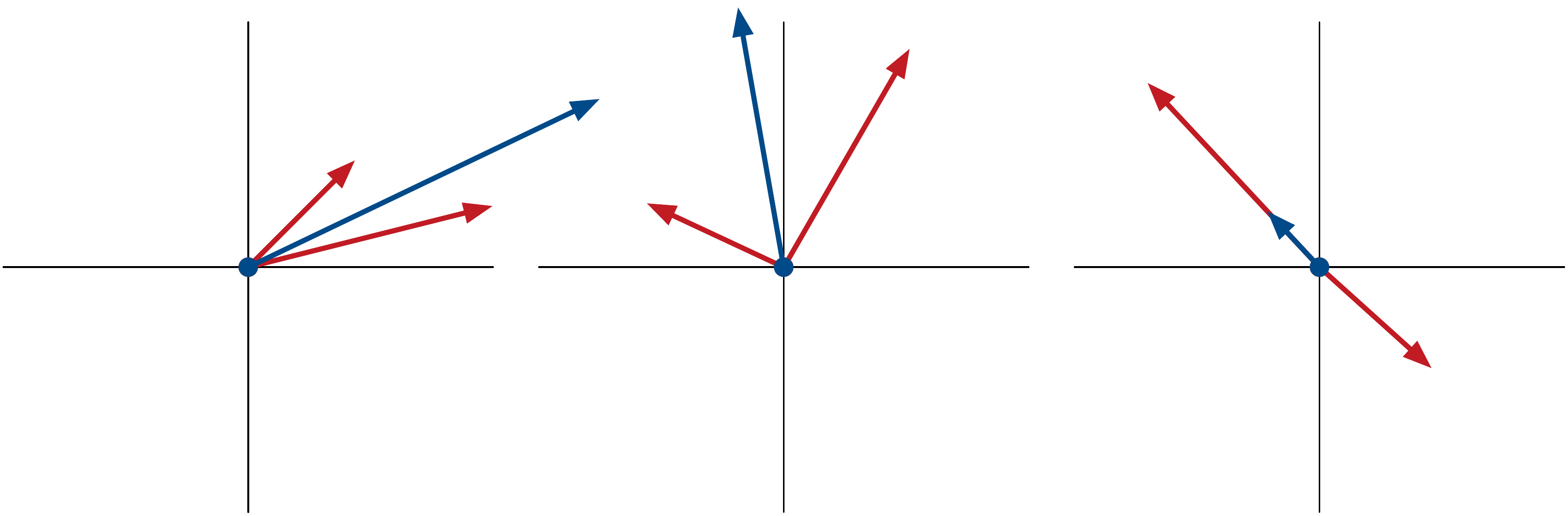}
\caption{Illustration of the change in direction and magnitude in the (blue) observed complex vector ${B^c_{\cal R}}+j{B^c_{\cal I}}$ created by the superposition of (red) complex vectors from multi-path fore and background objects for $K=1$, $K=8$, and $K=12$.}
\label{fig00}
\end{figure}

Now notice that by increasing the spatial frequency of the PMP patterns by a factor of $K$ increases the phase term by an equal amount while keeping the amplitude of the sinusoid constant.  In the case of multi-path, this frequency scaling has a far different affect as illustrated graphically in Fig.~\ref{fig00} where we show (left) the fore and background components assuming unit frequency while (center) and (right) show the same components when $K=8$ and $12$.  What Fig.~\ref{fig00} (left) shows in red are the complex vectors formed by $B^{c,f}_{\cal R}$ and $B^{c,f}_{\cal I}$ and $B^{c,b}_{\cal R}$ and $B^{c,b}_{\cal I}$, while the blue vector shows the superimposed vectors forming the single vector formed by $B^c_{\cal R}$ and $B^c_{\cal I}$. 

By using a frequency scaling of $K$, we expect the direction or phase of the foreground and background vectors to scale by an equal amount.  
Graphically, this is depicted by a rotation of the vectors around the origin.  Notice, though, that by rotating the vectors separately, that it is quite likely that the phase of the combined vectors are not equal to the scaling of the phase term prior to frequency scaling.  Likewise, the vectors may swing from constructively interfering where magnitude of the combine vectors is equal to the sum of the individual magnitudes to destructively interfering where the magnitude of the combine vectors is equal to the difference of the individual magnitudes. 

\begin{figure}[!t]
\centering\includegraphics[width=3.25in]{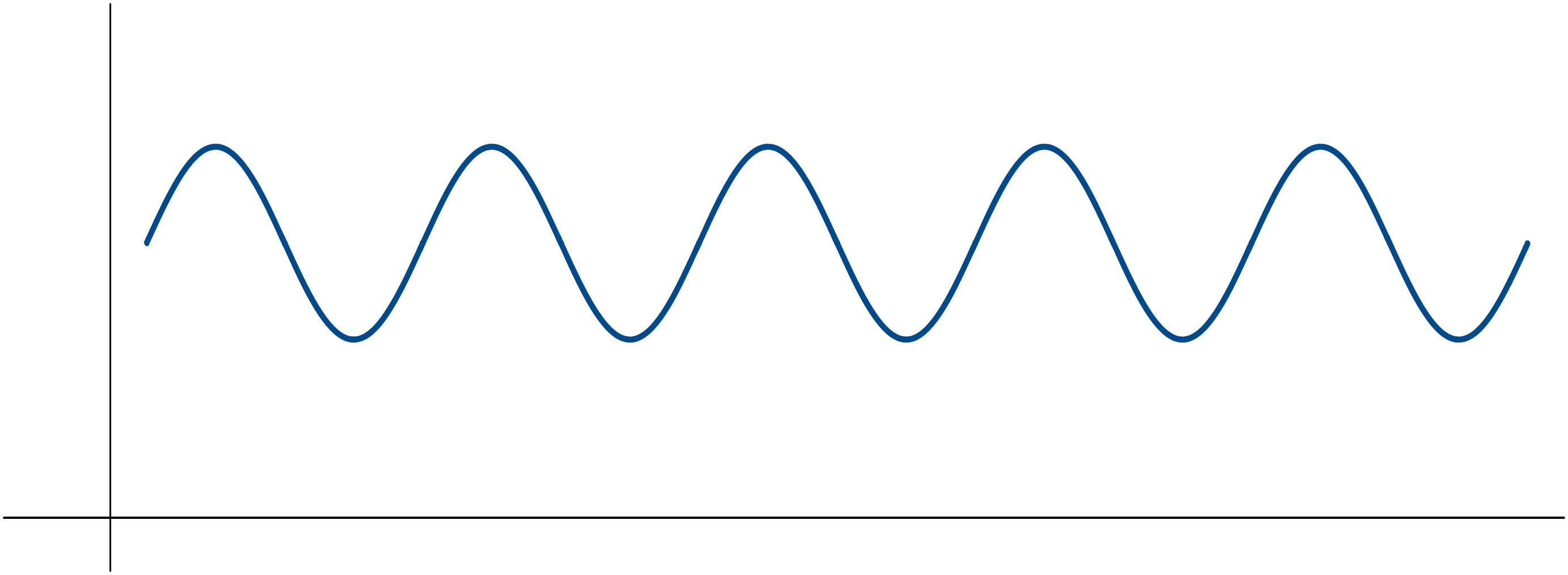}
\caption{Illustration of the change in magnitude in the observed complex vector ${B^c_{\cal R}}+j{B^c_{\cal I}}$ as a function of the scaling factor $K$.}
\label{fig01}
\end{figure}

\section{Bimodal Multi-Path Reconstruction}
\noindent Mathematically, the magnitude and phase of the subject pixel can be defined according to vector $\vec{AB}$  with foreground vector $\vec{A}$ and background vector $\vec{B}$ such that:
\begin{equation}
\label{EQ:BimodalMultipath}
|\vec{AB}|^2  =  |\vec{A}|^2 + |\vec{B}|^2 + 2 |\vec{A}||\vec{B}| cos(2\pi K(y^p_a - y^p_b))
\end{equation}
where $y_a$ and $y_b$ are the projector row coordinates for the two paths. It is this change in vector phase and magnitude in the superimposed vectors as a function of $K$ that is the prime means by which to detect multi-path in the scanned image. To separate the vectors $\vec{A}$ and $\vec{B}$ from $\vec{AB}$, a two-step procedure first finds the parameters $|\vec{A}|$, $|\vec{B}|$, and $dy = y^p_a - y^p_b$ that minimize the mean-squared error between $|\vec{AB}|$ and $|\vec{A} + \vec{B}|$ over all $K$, and then obtains the absolute phases $y^p_a$ and $y^p_b$ by minimizing the mean squared error between $\vec{AB}$ and $\vec{A} + \vec{B}$ with the constraints on $|\vec{A}|$, $|\vec{B}|$, and $dy$. We formulate it as eqn.~(\ref{EQ:solver}). 
\begin{equation}
\label{EQ:solver}
\arg\min_{|\vec{A}|,|\vec{B}|,dy} \sum_K \{ |\vec{AB}| - |\vec{A} + \vec{B}| \}^2
\end{equation}
To minimize the search space from three independent variables $|\vec{A}|$, $|\vec{B}|$, and $dy$ to the two $|\vec{A}|$ and $dy$, we define a zero-frequency scan where $K=0$ to obtain $\vec{AB}_{0}$ such that:
\begin{equation}
|\vec{AB}_{0}|^2  =  |\vec{A}|^2 + |\vec{B}|^2 + 2 |\vec{A}||\vec{B}|.
\end{equation}
From this, we get the constraint:
\begin{equation}
|\vec{A}| + |\vec{B}| = |\vec{AB}_{0}|
\end{equation}
so that we can perform an exhaustive search over $|\vec{A}|$, $|\vec{B}|$, and the phase difference $y^p_a - y^p_b$ along the line $|\vec{B}| = |\vec{AB}_{0}| - |\vec{A}|$ to find the values that minimize the mean-squared error in eqn.~(\ref{EQ:BimodalMultipath}) over all scanned values of $K$.

\begin{figure}[!t]
\centering\includegraphics[width=3.25in]{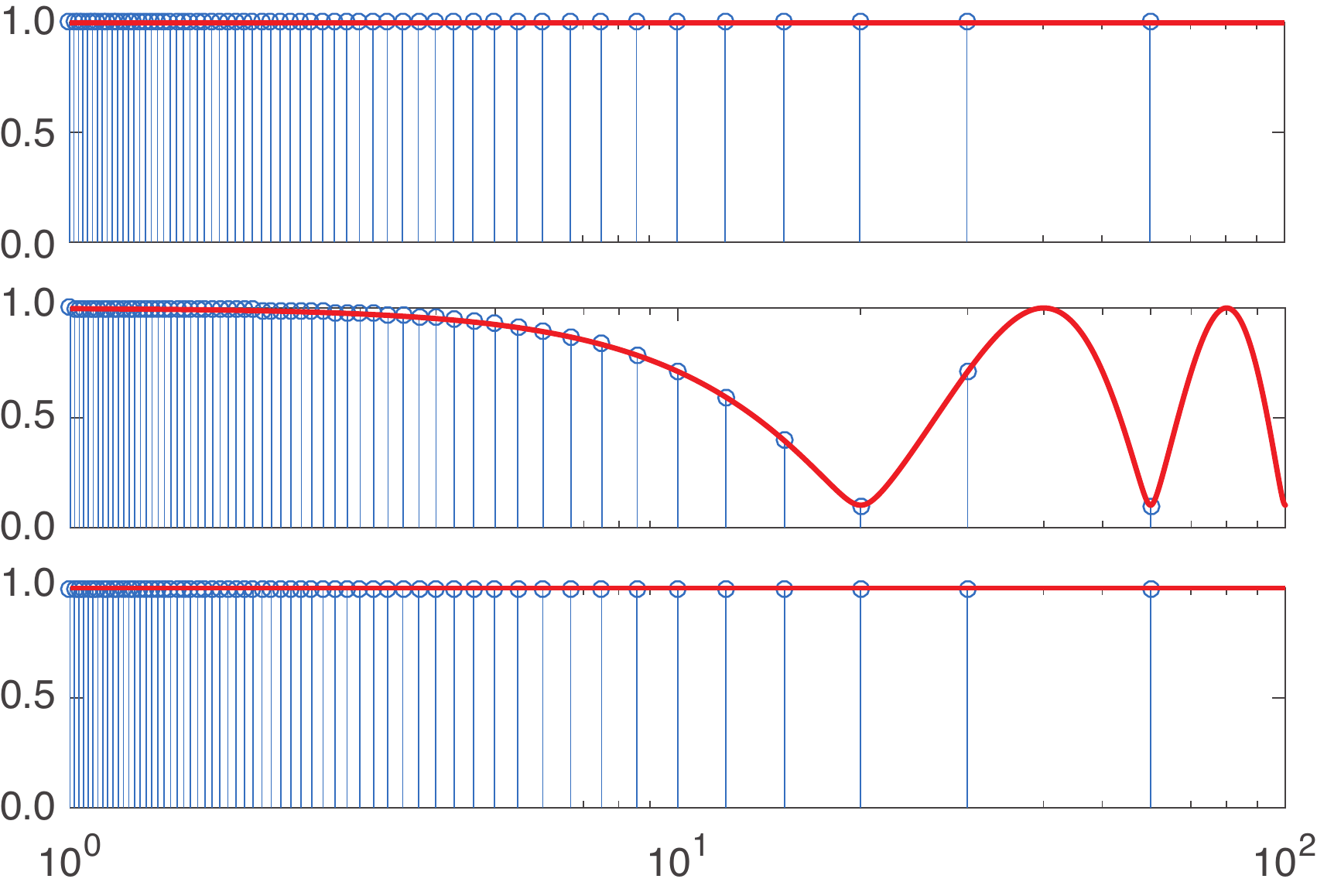}
\caption{Plots showing the ideal $|\vec{AB}|$ over $K$ for (top) the background pixel, (center) the edge pixel, and (bottom) the foreground pixel where the red line illustrates $|\vec{AB}|$ over continuous frequency.}
\label{fig05}
\end{figure}

As an illustration of the proposed algorithm, Fig.~\ref{fig05} shows plots of simulated $|\vec{AB}|$ over $K$ for two pixels, separated in the projector by 12 pixels, with (top) just the background pixel, (bottom) just the foreground pixel, and (center) a linear combination of 55\% foreground and 45\% background pixel.  As will be the case for these stem plots in this paper, the frequency, $K$, ranges from 1 to 60 sinusoids, at wavelength intervals of 8 pixels, across the projector field of view and is plotted in Fig.~\ref{fig05} on the log scale.  Also note that the y-axis is normalized by $|\vec{AB}_{0}|$ and will range from 0 to 1. Shown in red are plots of the resulting best-fit $\vec{A}$ and $\vec{B}$ vectors where $|\vec{AB}|$ is plotted over continuous $K$ from 1 to 60 Hz.


\section{Experimental Evaluations}
\begin{figure}[!t]
\centering\includegraphics[width=3.25in]{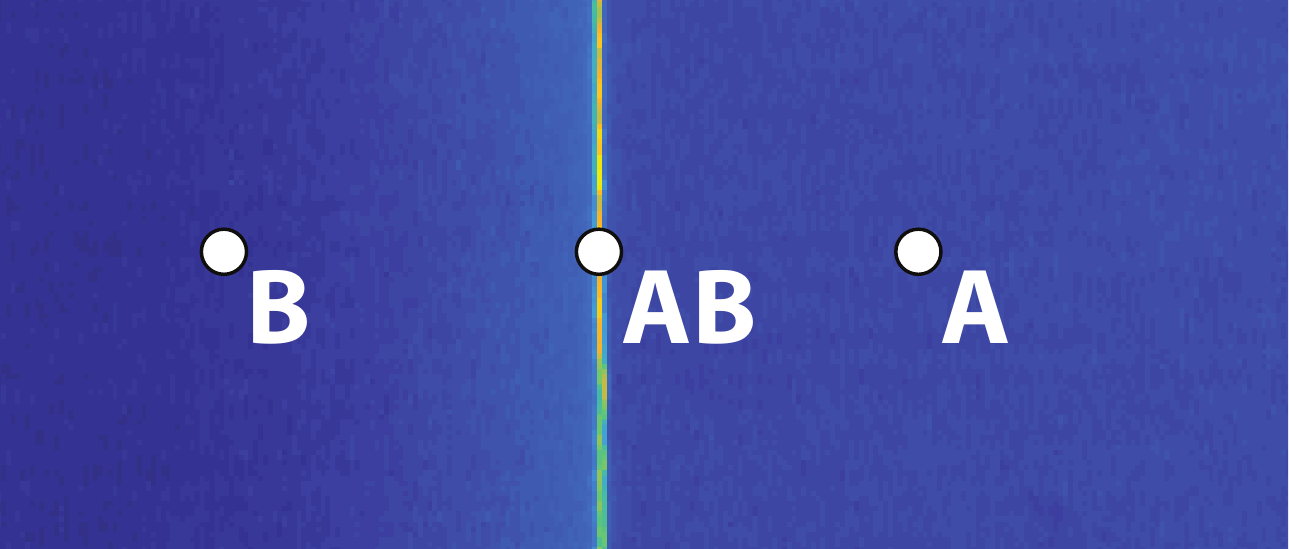}
\caption{Illustration of the variance in $|\vec{AB}|$ over $K$ for a textureless surface with a step edge with a foreground pixel labeled ${\bf A}$, a background pixel labeled ${\bf B}$, and edge pixel ${\bf AB}$.}
\label{fig02}
\end{figure}

\begin{figure}[!t]
\centering\includegraphics[width=3.25in]{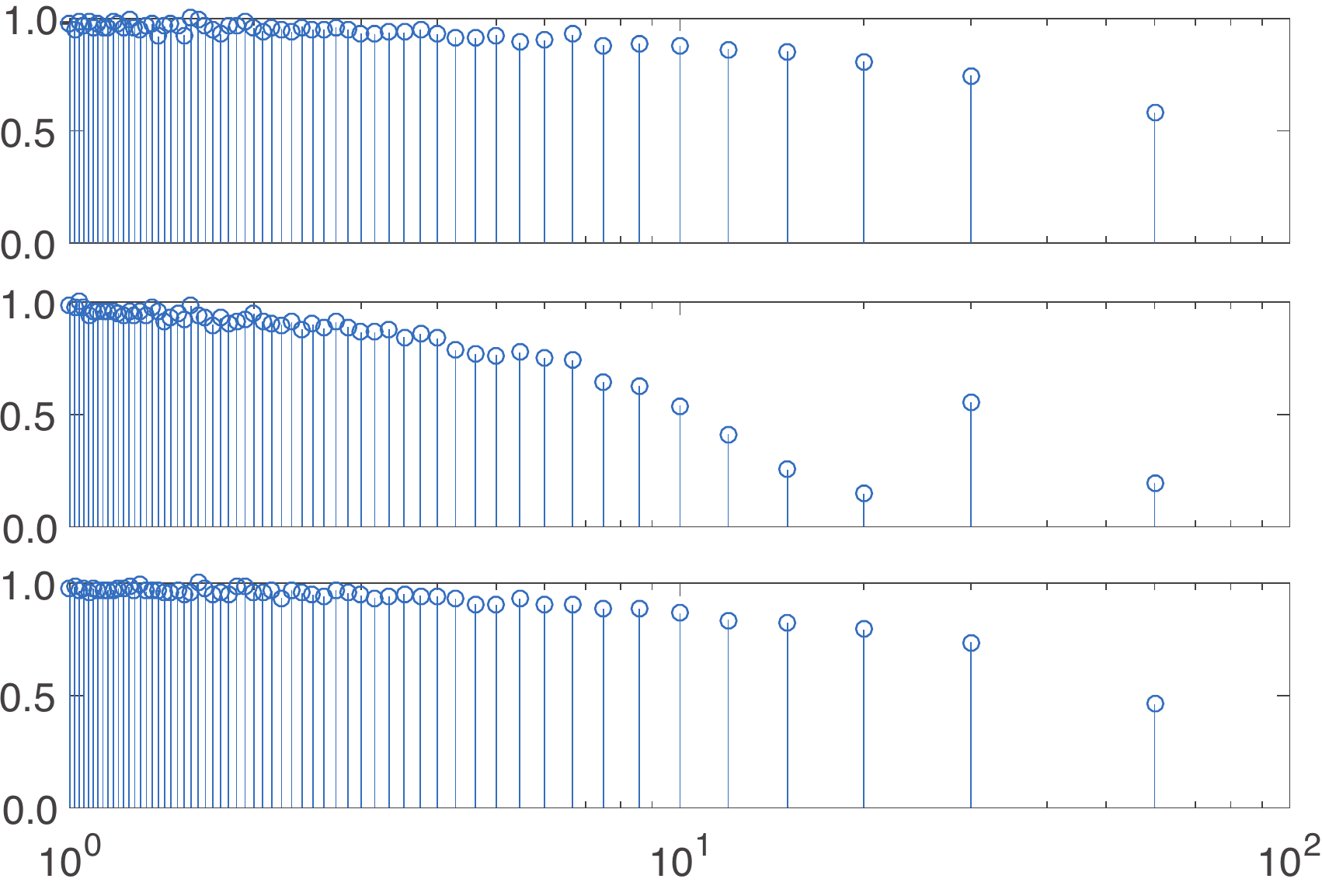}
\caption{Stem plots showing the measured $|\vec{AB}|$ over $K$ for (top) the background pixel ${\bf B}$, (center) the edge pixel ${\bf AB}$, and (bottom) the foreground pixel ${\bf A}$.}
\label{fig03}
\end{figure}

\noindent In order to demonstrate the proposed de-coupling technique, we consider the case of scanning two layers of half-inch, textureless, foam board where Fig.~\ref{fig02} shows the variance in the magnitude in the observed phasors, $\vec{AB}$, over all $K$ where the step edge is clearly visible as indicated by the bright vertical line.  To illustrate this sinusoidal shape on $|\vec{AB}|$, Fig.~\ref{fig03} shows stem plots of $|\vec{AB}|$ versus $K$ for the three pixels of Fig.~\ref{fig02}, labeled $A$, $B$, and $AB$ where $A$ corresponds to the foreground surface to the right of the edge, $B$ the background surface to the left of the edge, and $AB$ a pixel on the edge of the surface.  

\begin{figure}[!b]
\centering\includegraphics[width=3.25in]{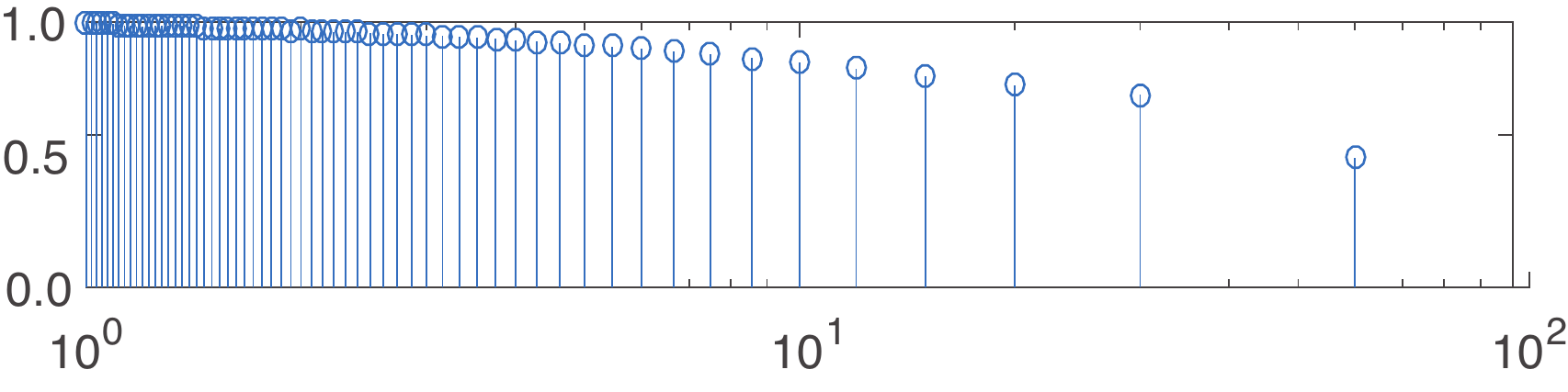}
\caption{Stem plot of the measured $|\vec{AB}|$ over $K$ for a flat, textureless surface at the center of the scanners focal distance averaged over all pixels as an estimate of the systems modulation transfer function.}
\label{fig04}
\end{figure}

\begin{figure}[!t]
\centering\includegraphics[width=3.25in]{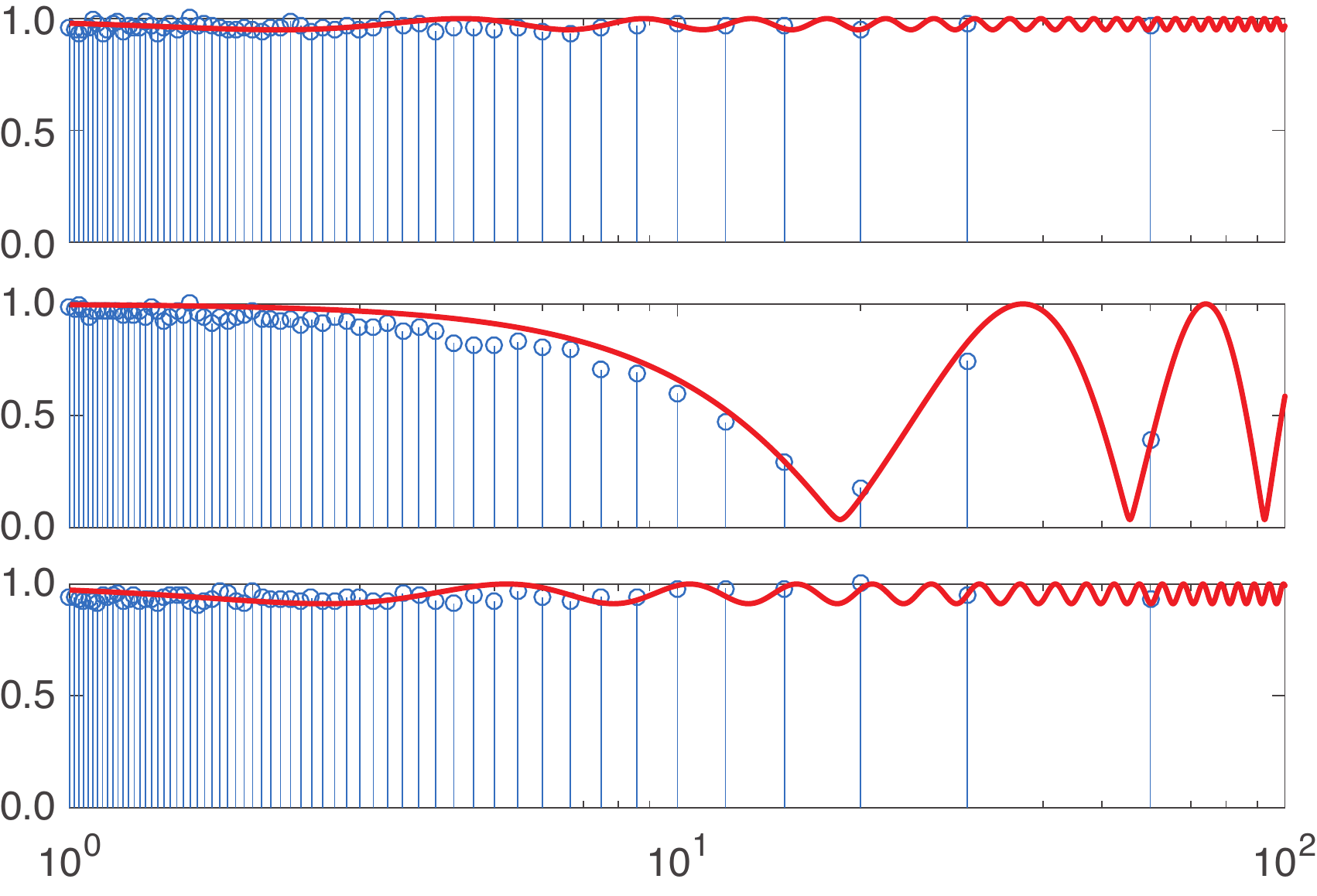}
\caption{Plots showing the measured $|\vec{AB}|$ over $K$ for (top) the background pixel ${\bf B}$, (center) the edge pixel ${\bf AB}$, and (bottom) the foreground pixel ${\bf A}$ where the red line illustrates best-fit $|\vec{AB}|$ over continuous frequency. }
\label{fig06}
\end{figure}

Observing the stem plot of Fig.~\ref{fig03}, one can see a consistent drop in magnitude at higher frequencies.  This is caused by the low-pass nature of the projector and camera optics, blurring the peaks and valleys of the projected sinusoids.  In order to account for the modulation transfer function of the projector/camera optics, we scan a white, textureless foam board at the center of our depth range and then average the value of $|\vec{AB}|$ over all pixels for all $K$ to produce the stem plot of Fig.~\ref{fig04}.  This resulting vector is then used as a normalizing factor for all subsequent scans. Applying this normalization to Fig.~\ref{fig03} produces the stem plots of Fig.~\ref{fig06} which now shows the expected flat response to fore and background pixels $A$ and $B$ and the distinctive sinusoidal shape for the edge pixel $AB$. 

Using the proposed algorithm on the edge pixel ${\bf AB}$, we obtained the normalized magnitudes of 0.5560 and 0.4440 and phases of 0.3647 and 0.3917 (projector row coordinates 188 and 175 or 13 pixels difference), respectively, where the actual pixels have phases of 0.3650 and 0.3916.  If we also apply the algorithm to background pixel ${\bf B}$ under the assumption of multi-path, we extract magnitudes of 0.9770 and 0.0230 with phase values 0.3918 and 0.1698, resulting in the small sinusoidal curve.  For the foreground pixel ${\bf A}$, we extract magnitudes of 0.9680 and 0.0320 with phase values 0.3648 and 0.0998.  We associate these weak, secondary multi-path signals to noise in the sensor and, ignoring these terms, focus on the edge pixel, ${\bf AB}$, noting how close our estimated values are to the true phases derived through the traditional structured light phase processing.

\begin{figure}[!t]
\centering\includegraphics[width=3.25in]{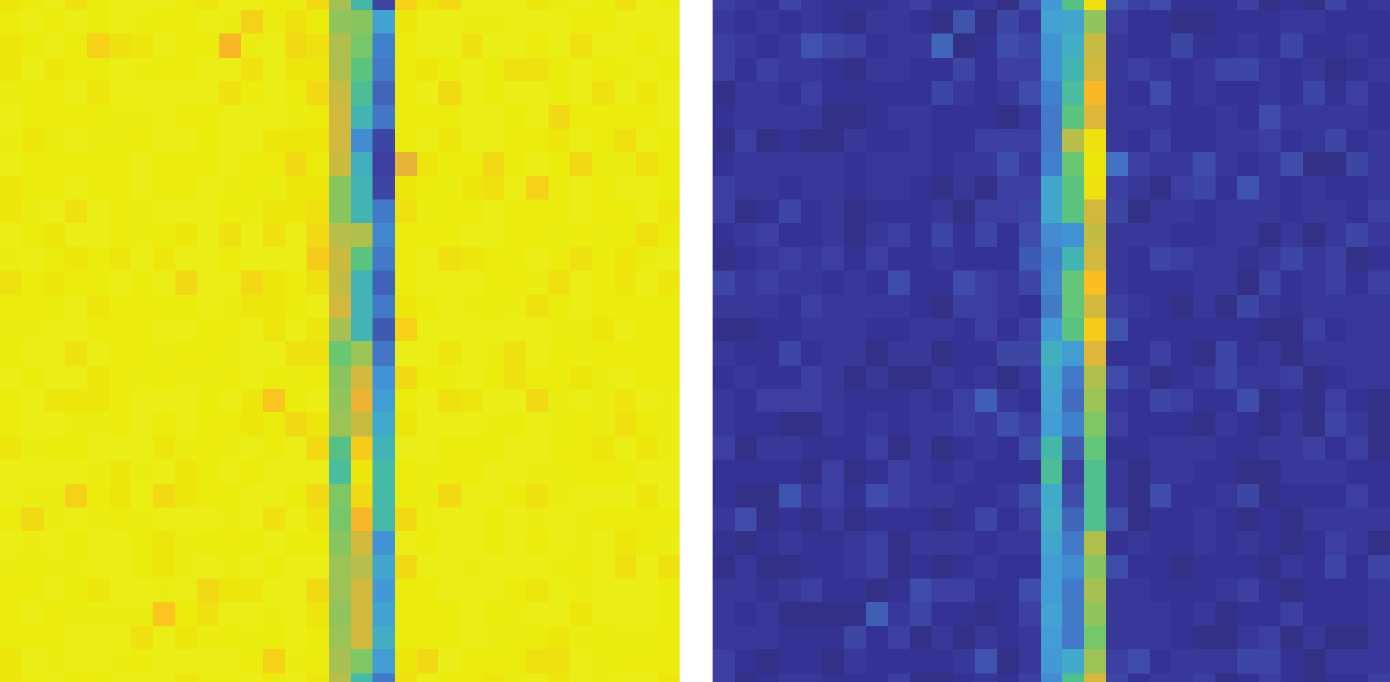}
\caption{Pseudo-color plot of the magnitudes of the primary (stronger) and secondary (weaker) bimodal path component along the step edge of Fig.~\ref{fig02}.}
\label{fig07}
\end{figure}

Applying the exhaustive search over $|\vec{A}|$, $|\vec{B}|$, and the phase difference $y^p_a - y^p_b$ along the line $|\vec{A}|^2 + |\vec{B}|^2 = |\vec{AB}_{0}|^2$ for a small region of interest about the step edge. The exhaustive search for each pixel can be done within 12 seconds due to the limited searching space. With a GPU implementation, we can solve these pixels in parallel since each pixel is independent to others. Fig.~\ref{fig07} shows the value of the magnitude of the (left) primary, the larger of $|\vec{A}|$ or $|\vec{B}|$, and the (right) secondary or smaller term.  The corresponding primary and secondary phase terms are illustrated in Fig.~\ref{fig08}. Relying on the primary term for reconstructing depth, Fig.~\ref{fig08} illustrates the improved edge rendition sans bimodal multipath.

\begin{figure}[!t]
\centering\includegraphics[width=3.25in]{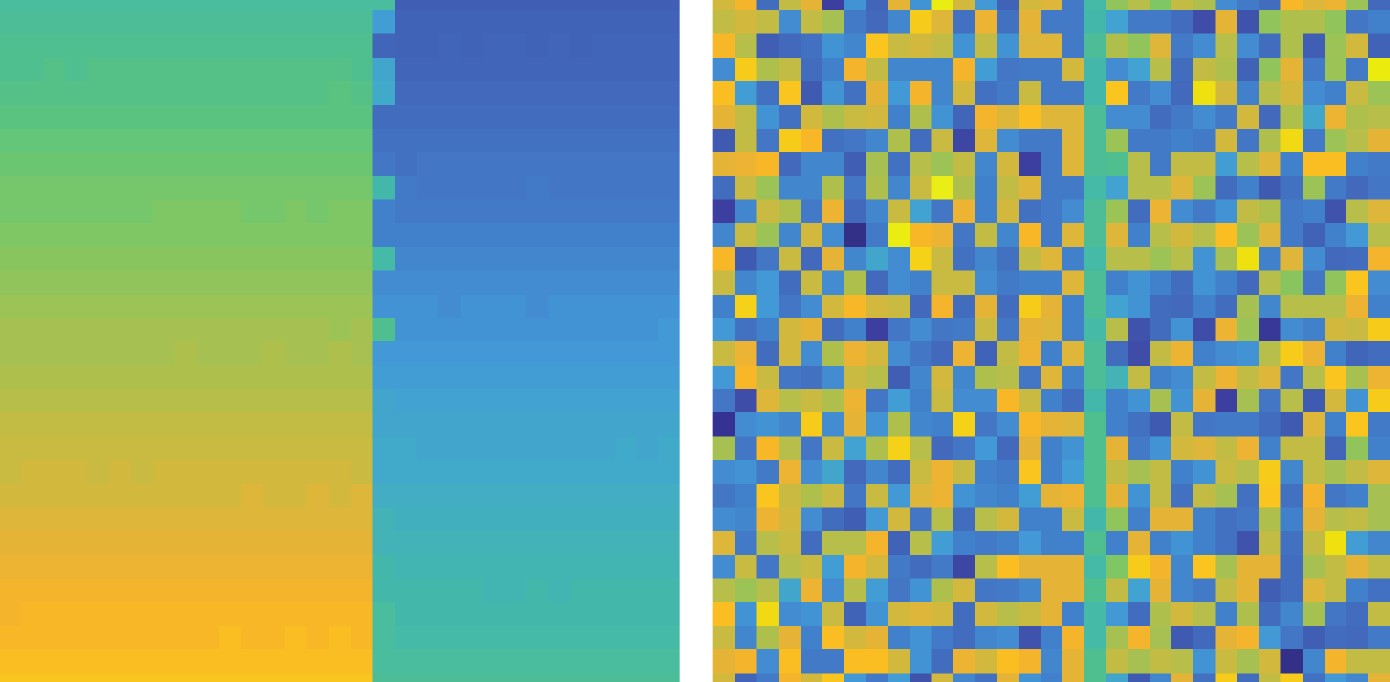}
\caption{Pseudo-color plot of the phases of the primary (stronger) and secondary (weaker) bimodal path component along the step edge of Fig.~\ref{fig02}.}
\label{fig08}
\end{figure}

\begin{figure}[!t]
\centering\includegraphics[width=3.25in]{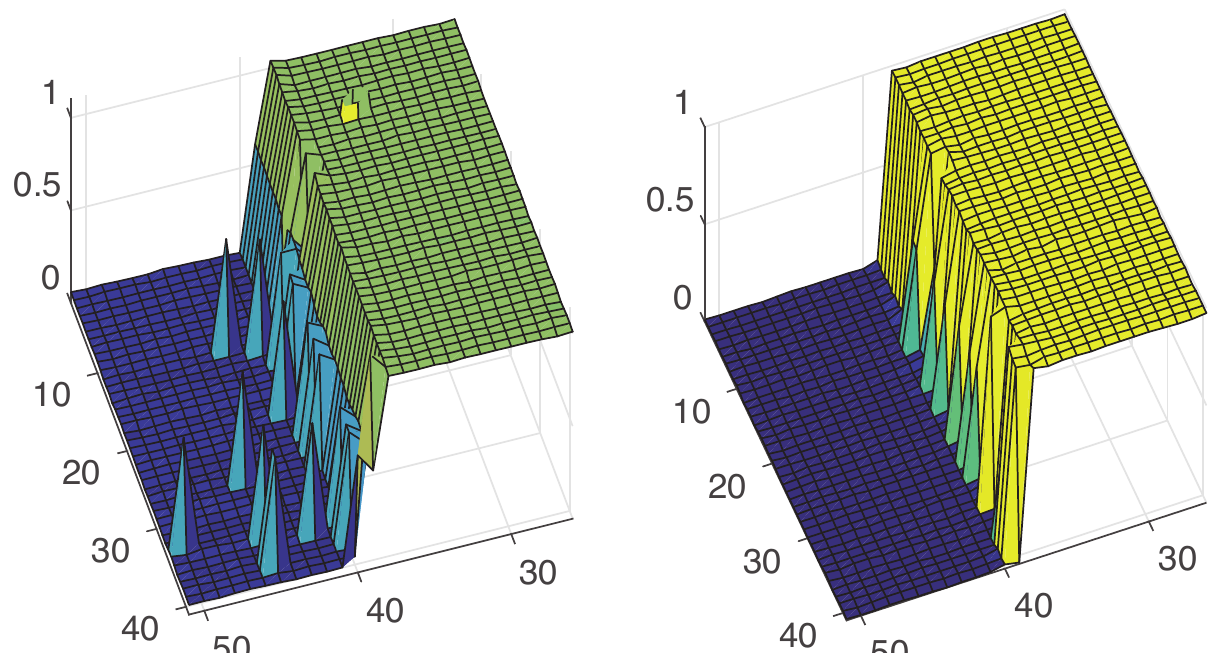}
\caption{Surface plots of the phase images (left) before and (right) after applying the multi-path separation procedure where (left) shows the unprocessing phase image while (right) shows the phase image of the dominant component from Fig.~\label{fig08}.}
\label{fig09}
\end{figure}

\begin{figure}[!t]
	\centering\includegraphics[width=3.25in]{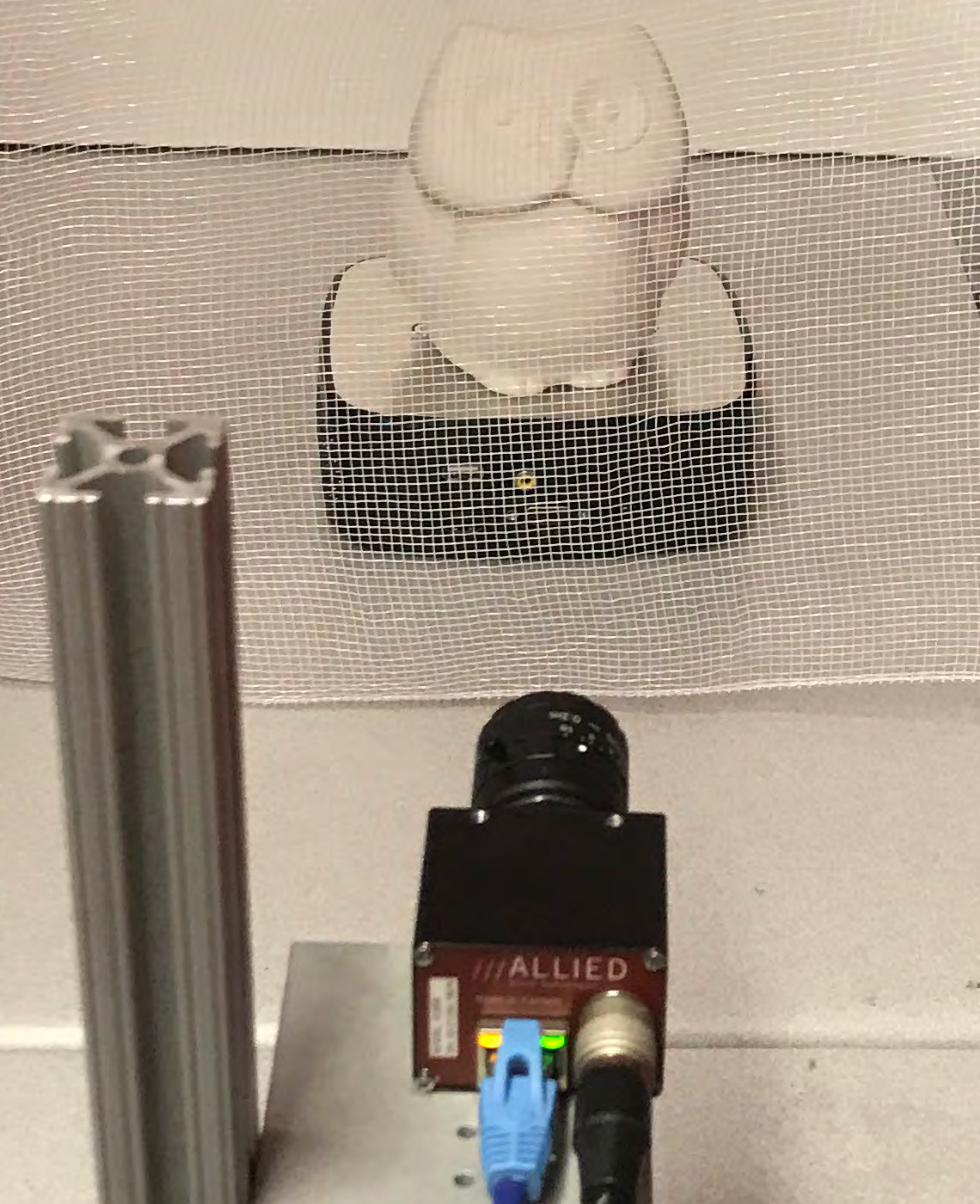}
	\caption{Experimental setup scanning a white owl figurine through a polyester cloth mesh.}
	\label{img00}
\end{figure}

\begin{figure}[!t]
	\centering\includegraphics[width=3.25in]{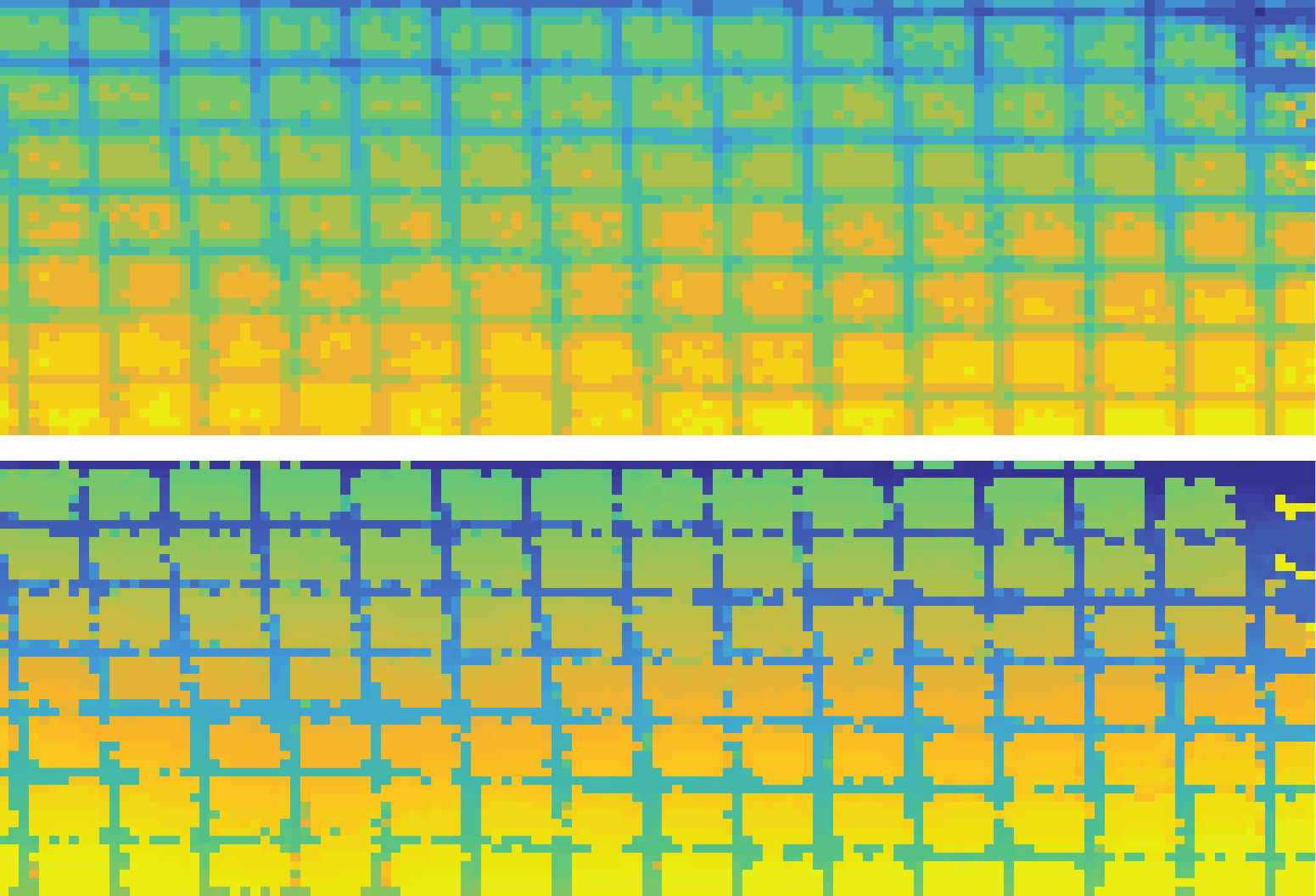}
	\caption{Pseudo-color plot of the phases of the (top) raw phase image of the owl figurine beak and eyes through the mesh and the (bottom) primary  bimodal path component.}
	\label{fig11}
\end{figure}


\begin{figure}[t]
	\centering\includegraphics[width=3.45in]{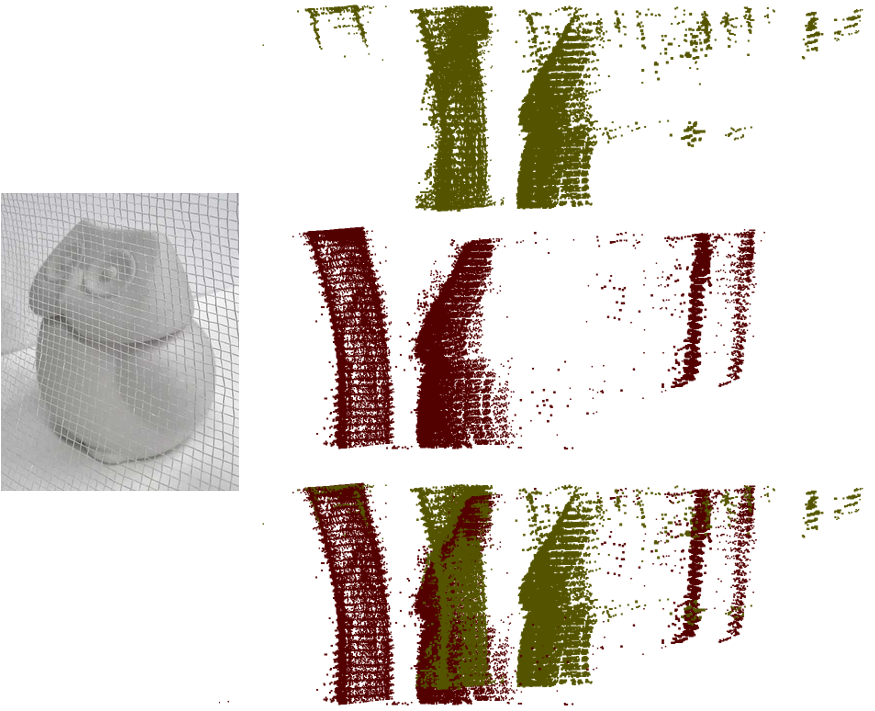}
	\caption{Point cloud reconstructions of the owl figurine using raw phase (top), multi-path processed phase (middle) and both before and after in a single coordinate space (bottom).}
	\label{owl}
\end{figure}

For a demonstration of multi-path separation in a structured light system, Fig.~\ref{img00} shows an experimental setup where we scanned a white plaster owl figurine through a polyester cloth mesh.  Shown in Figs.~\ref{fig11} and \ref{owl} are the resulting phase reconstructions and point cloud showing the before effects of using the proposed multi-path detection scheme. It should be evident that this phase unwrapping error is devastating to the 3D reconstruction which we illustrate in Fig.~\ref{owl} where the reconstruction from the raw phase image is given in Fig.~\ref{owl}~(top) while Fig.~\ref{owl}~(middle) shows the reconstruction using the multi-path phase image.  Its not a mistake that the raw phase image produces a reconstruction that is shifted in Z by 50 millimeters. What is especially fascinating is placing both reconstructions in a common coordinate system is illustrated in Fig.~\ref{owl} ~(bottom) where there are phantom dots in the raw phase reconstruction that do perfectly correspond to points in the multi-path reconstruction.  Again, the multi-path reconstructions are correct, the traditional phase unwrapping is incorrect in these figures. From visual inspection, the proposed technique is a clear improvement over the scan produced without the process.

\begin{figure}[!t]
	\centering\includegraphics[width=1.0in]{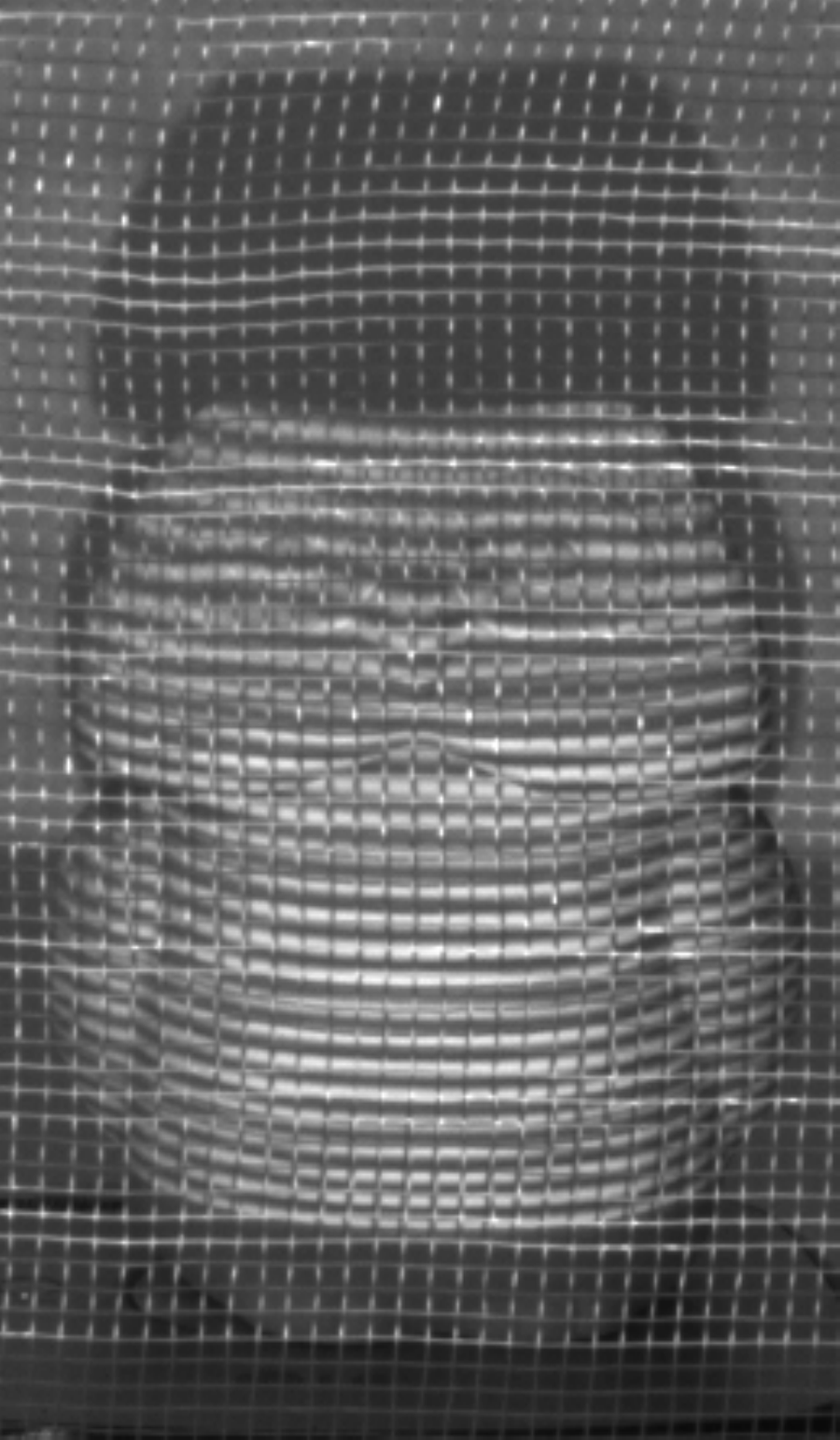}
	\caption{Raw video frames for the $f=60$ cosinusoidal grating pattern.}
	\label{Pab03}
\end{figure}

For a more detailed analysis of the reconstructions produced using the new multi-path procedure, note that shown in Fig.~\ref{Pab03} is the first of eight frames corresponding to a sinusoidal wavelength of 8 projector pixels with 60 wavelengths across the projected image from top to bottom. We specifically looked at pixel with row and column coordinate $[278,319]$ since it appears just under the owl's chin and perfectly situated between the threads of the foreground screen. At the same time, this pixel was also selected because it sits on the boundary between the 15th and 16th wavelengths.  As such, we know its true phase is equal to $\frac{16}{60} = 26.67\%$ of the projected phase range with 0 corresponding to the bottom of the projected image and 100\% corresponding to the top.

\begin{figure}[!t]
	\centering\includegraphics[width=0.90in]{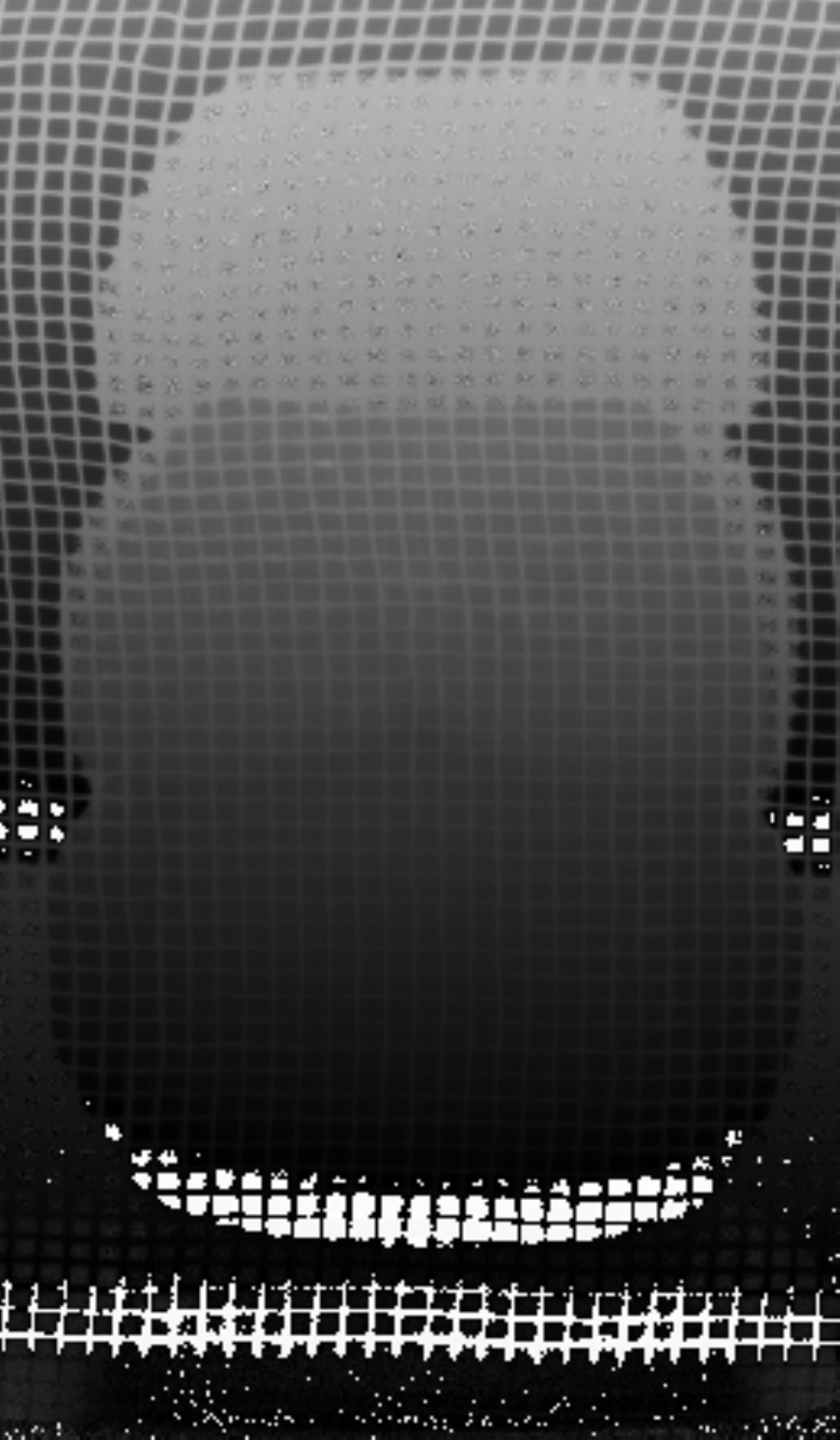} \hspace{0.05in}
	\centering\includegraphics[width=0.90in]{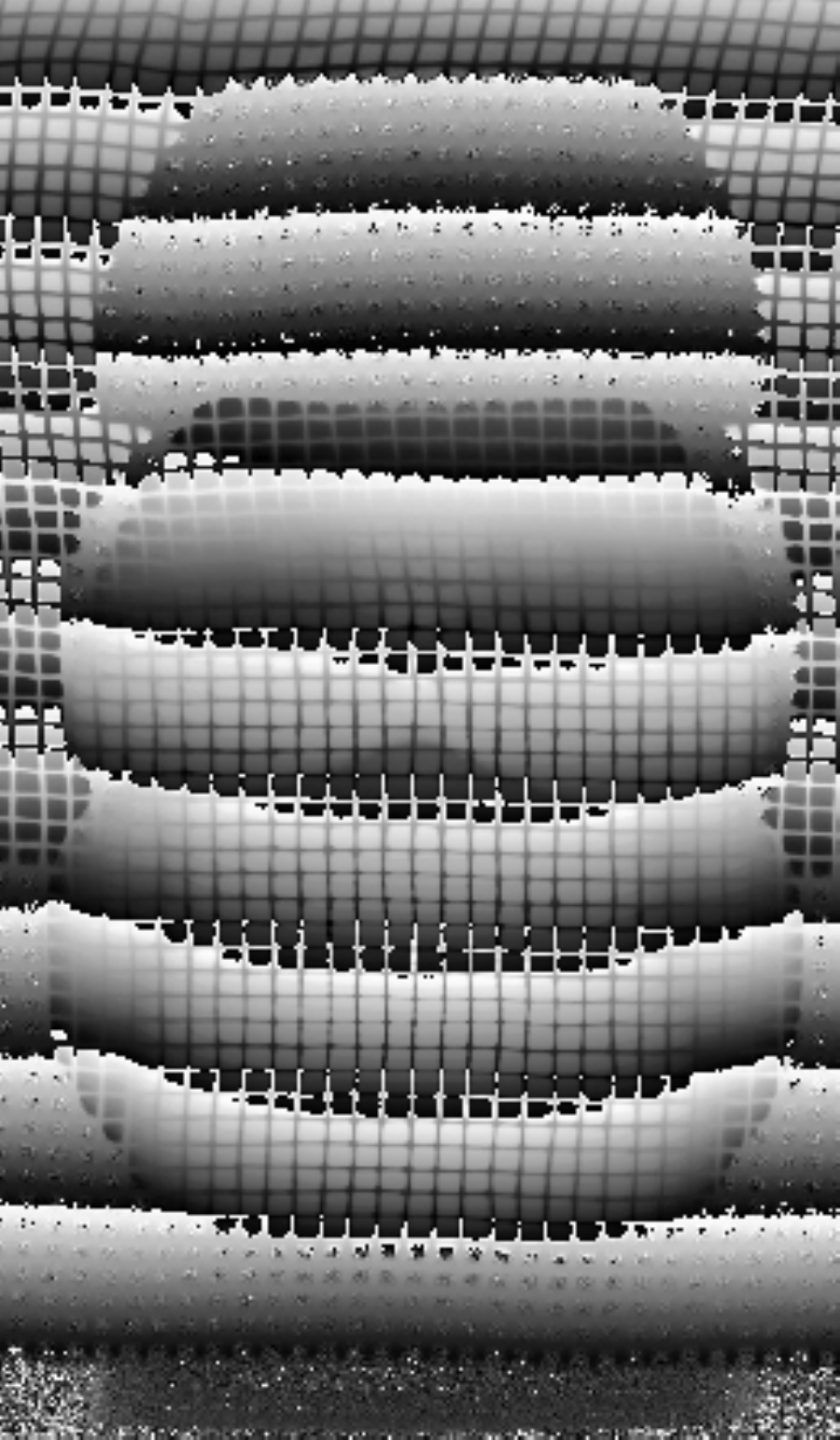} \hspace{0.05in}
	\centering\includegraphics[width=0.90in]{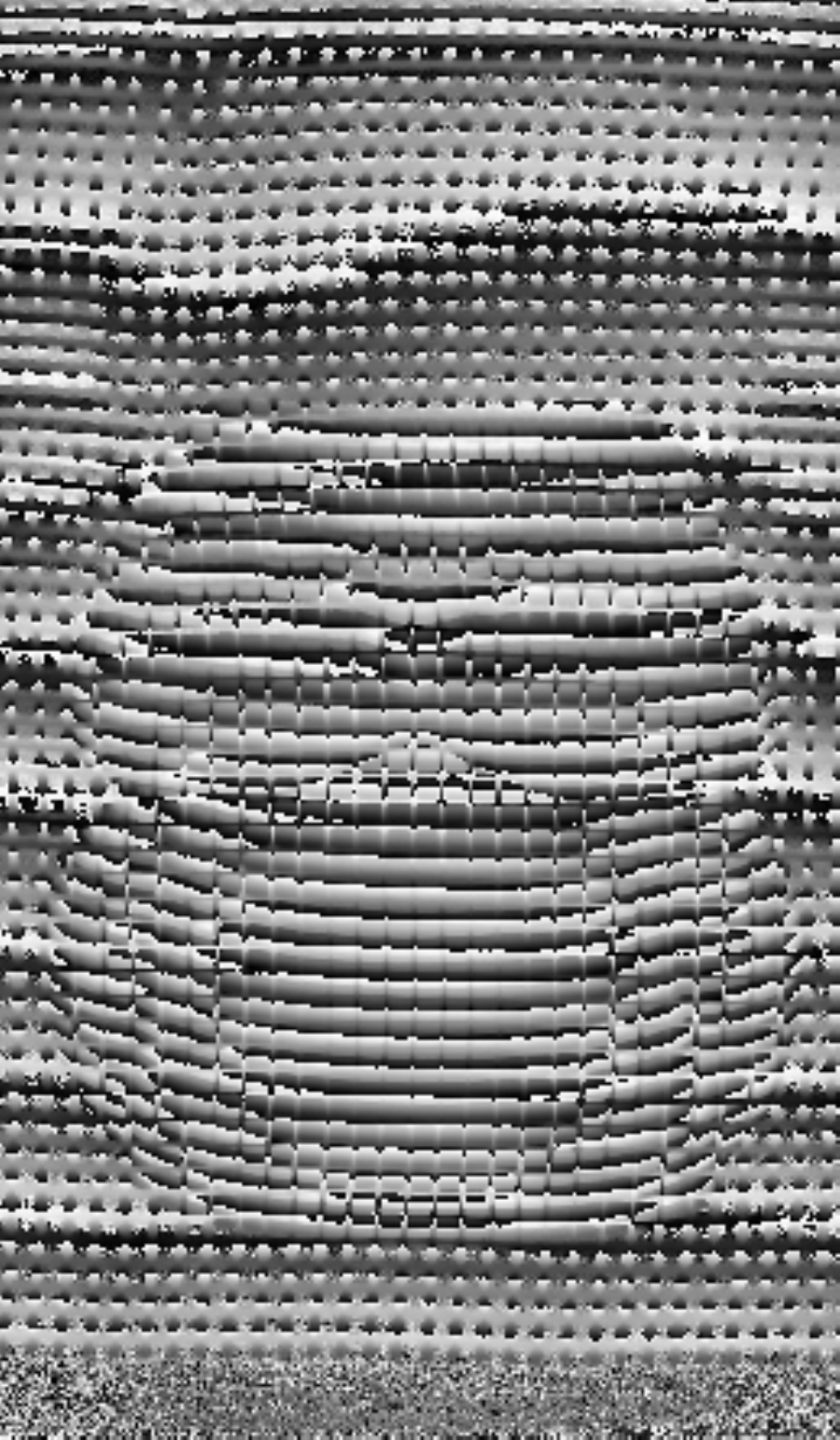}
	\caption{The unit, mid, and high frequency phase images for frequencies $f=1$, $12$, and $60$ sinusoidal gratings.}
	\label{figpah3}
\end{figure}

For generating the raw phase image, we employed the traditional procedure of using three unique pattern frequencies of 1, 12, and 60 wavelengths.  The 12 wavelengths were first unwrapped using the unit frequency pattern, and then this smooth unit frequency image was used to unwrap the 60 wavelengths.  Looking at pixel $[278,319]$, the traditional phase unwrapping process determined that the pixel had a phase of $18.27\%$, an error of $8.33\%$. Using the proposed multipath method produces a phase estimate of $26.76\%$, an insignificant error within round off of one pixel.  What this large error in traditional phase unwrapping can be attributed to is an error in the mid-frequency phase image, which incidentally would corresponding to plus or minus one wavelength or $\frac{1}{12}=8.33\%$.

To see this phase unwrapping error in action, Fig.~\ref{figpah3} shows the raw phase images for 1, 12, and 60 wavelengths where the foreground screen interacts with the backgrounds to create a moir\'e pattern, which is a low-frequency sinusoidal grating created by the superposition of higher frequency gratings.  Note most importantly that there is a swatch of phase values in the area directly underneath the owl figurine, where light from the projector only intersects the foreground screen since the bottom of the projected image first reflects off the figurine about 1/4-inch from its base.

\begin{figure}[!t]
	\centering\includegraphics[width=3.4in]{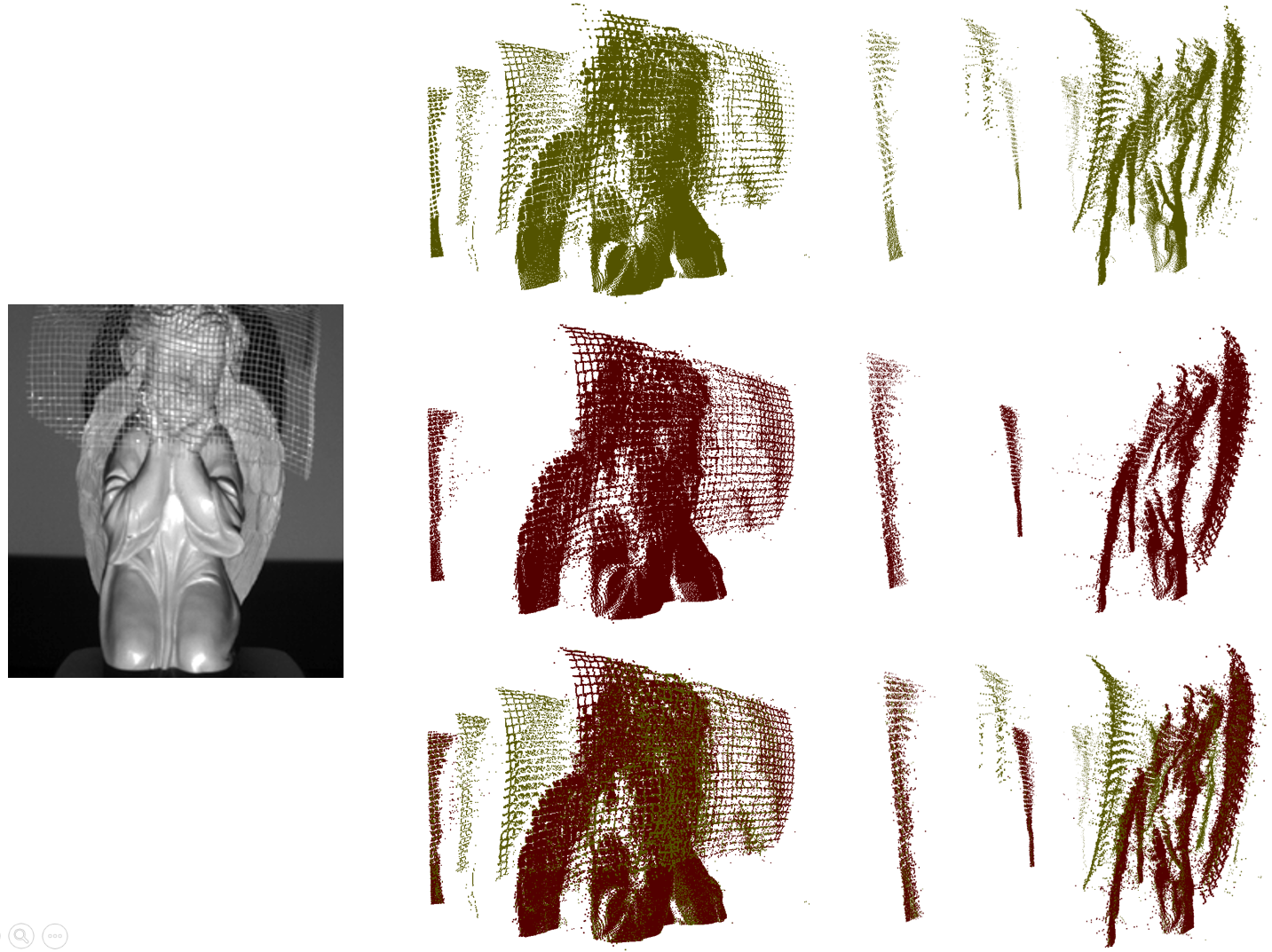}
	\caption{Point cloud reconstructions of the angel figurine (left) using raw phase (in yellow on the first row), multi-path processed phase (in red on the second row), and both before and after in a single coordinate space (in a mixture of red and yellow on the third row).}
	\label{angel}
\end{figure}

As a similar demonstration of multi-path reconstruction, Fig.~\ref{angel} shows an angel figurine with the same screen placed over the angel's head and shoulders. The 3D point cloud reconstructions are shown in yellow on the first row showing the raw phase reconstruction, in red on the second row showing the multi-path reconstruction, and in a mixture of red and yellow on the third row showing both in a single coordinate space. In this illustration, we note that without the multi-path algorithm, traditional SLI reconstruction will result in multiple ghost layers of the screen that appear at incorrect position in front and behind the figurine. With our proposed algorithm, the ghost layers disappear and result in an accurate reconstruction of the screen in front of the figurine.

\begin{figure*}[!t]
	\centering\includegraphics[width=1.80in]{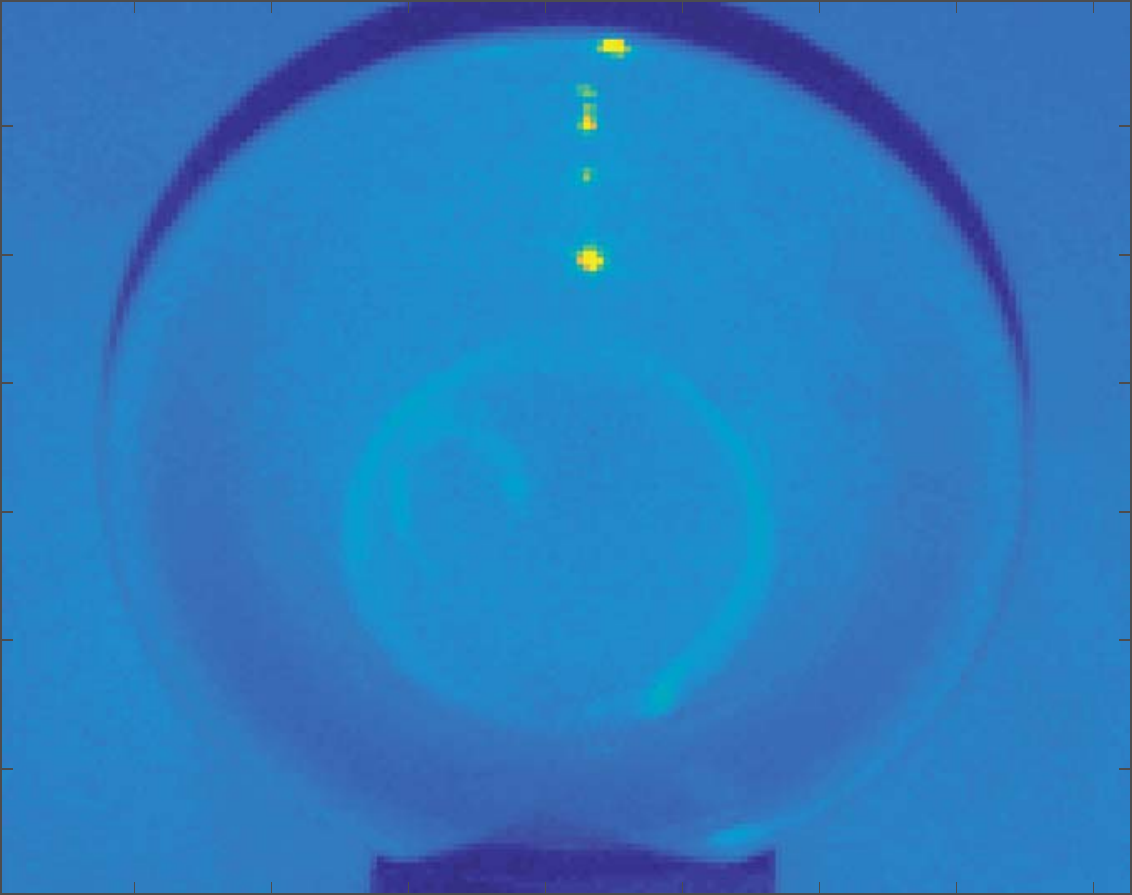}
	\centering\includegraphics[width=1.80in]{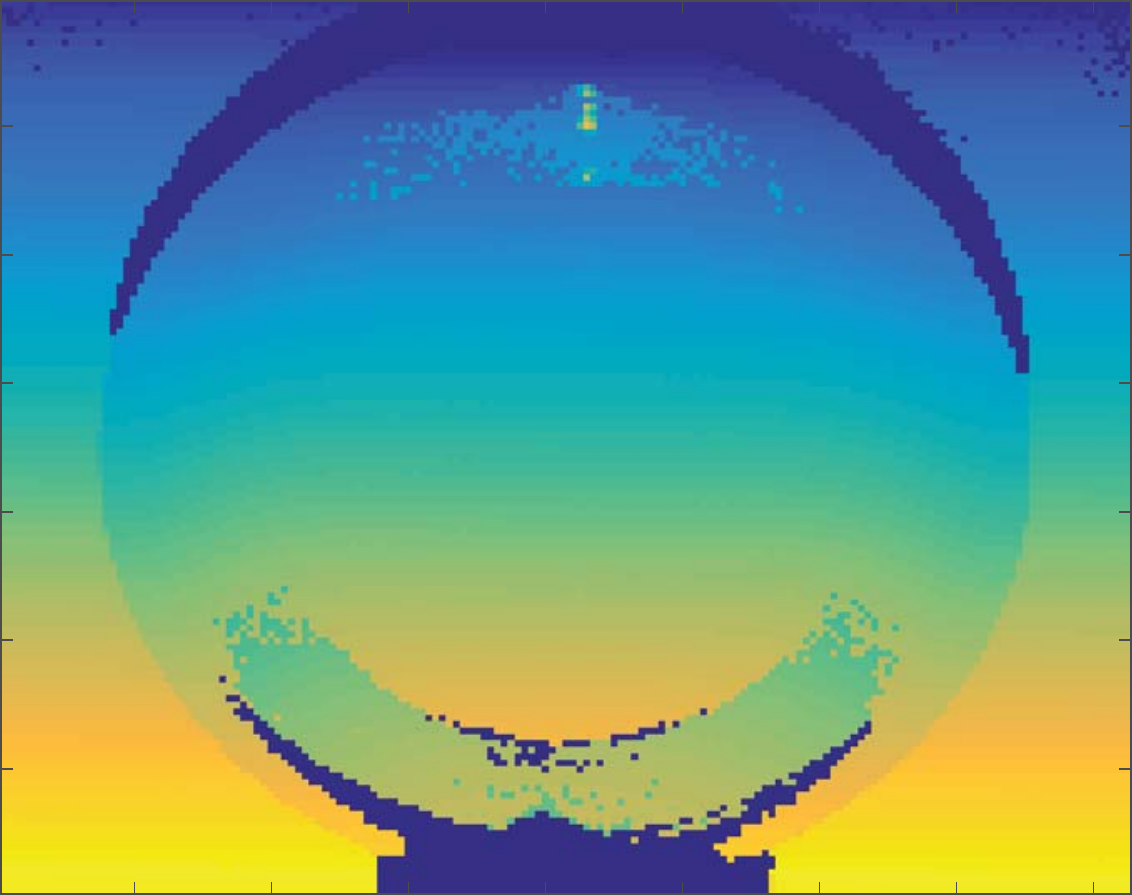}
	\centering\includegraphics[width=1.80in]{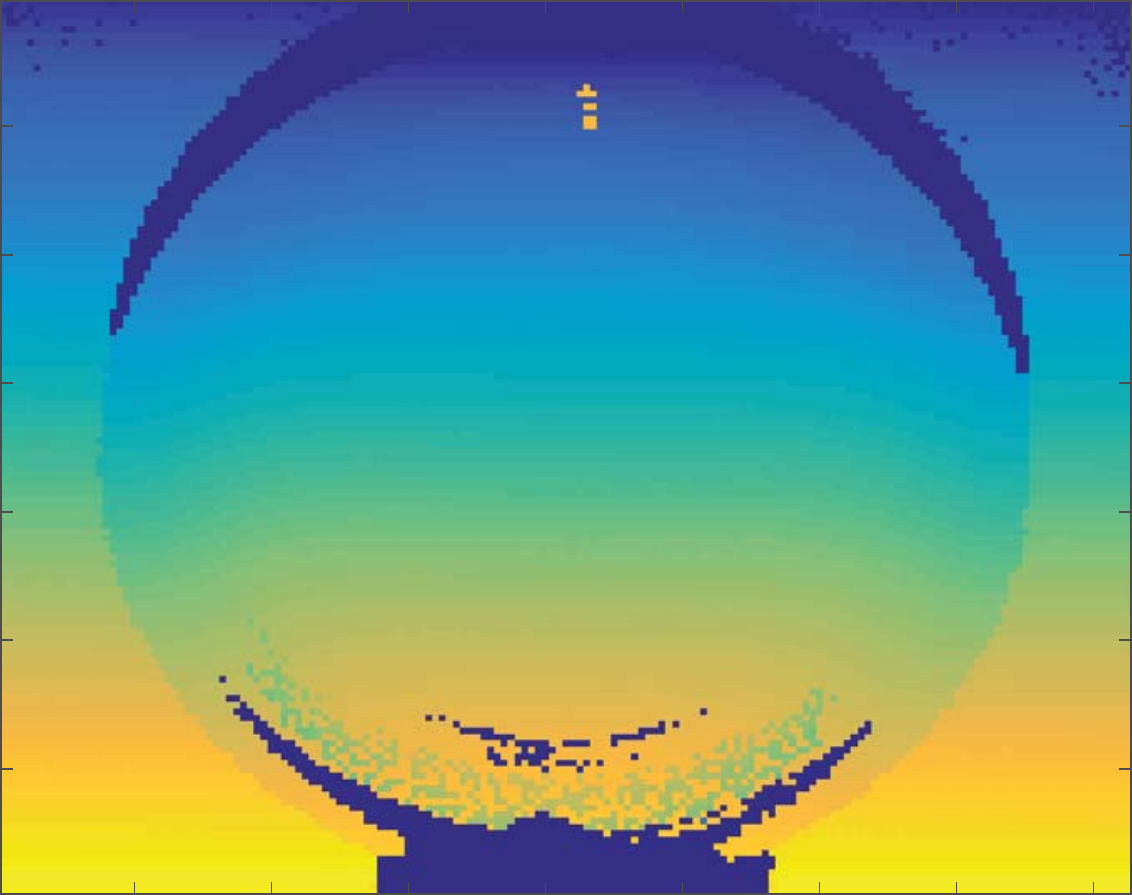}
	\caption{The (center) raw and (right) multi-path phase images derived from looking into a (left) porcelain bowl.}
	\label{bowl}
\end{figure*}

\begin{figure*}[!t]
	\centering\includegraphics[width=2.20in]{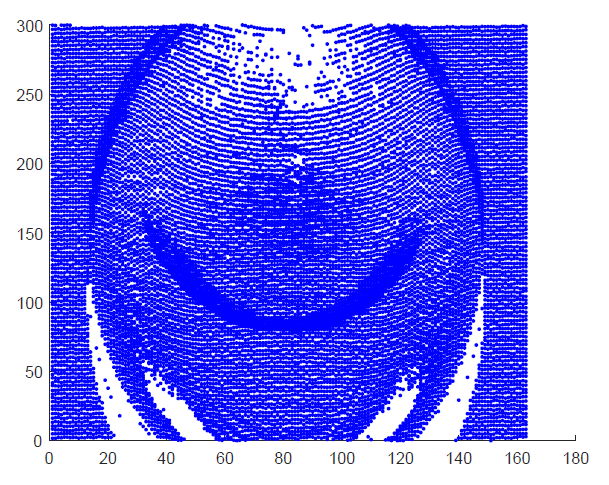}
	\centering\includegraphics[width=2.20in]{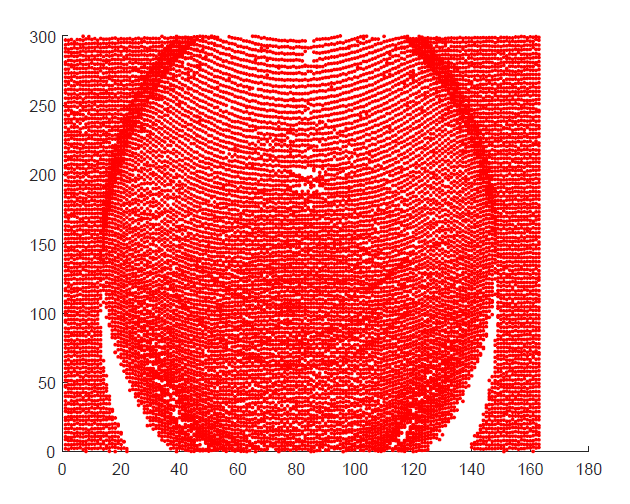}
	\caption{Point cloud reconstructions of the bowl using (left) raw phase and (right) multi-path processed phase.}
	\label{bowlresu}
\end{figure*}

As a third demonstration of the multi-path technique, Figs.~\ref{bowl} and \ref{bowlresu} show the phase and point cloud reconstructions comparing again the traditional phase unwrapping procedure versus our proposed multi-path procedure when the target image is the inside of a white, porcelain bowl.  In this sample, specular reflections off the surface of the bowl create multi-paths, most evident at the top and the bottom of the bowl where the reflections stay within the epipolar geometry of the camera/projector lens alignment.  While the new multi-path procedure is not completely immune to issues cause by specularities on the target surface, it is greatly improved over the board artifacts introduced through phase unwrapping, as indicated  in Fig.~\ref{bowlresu}.

\begin{figure}[t]
	\centering\includegraphics[width=1.60in]{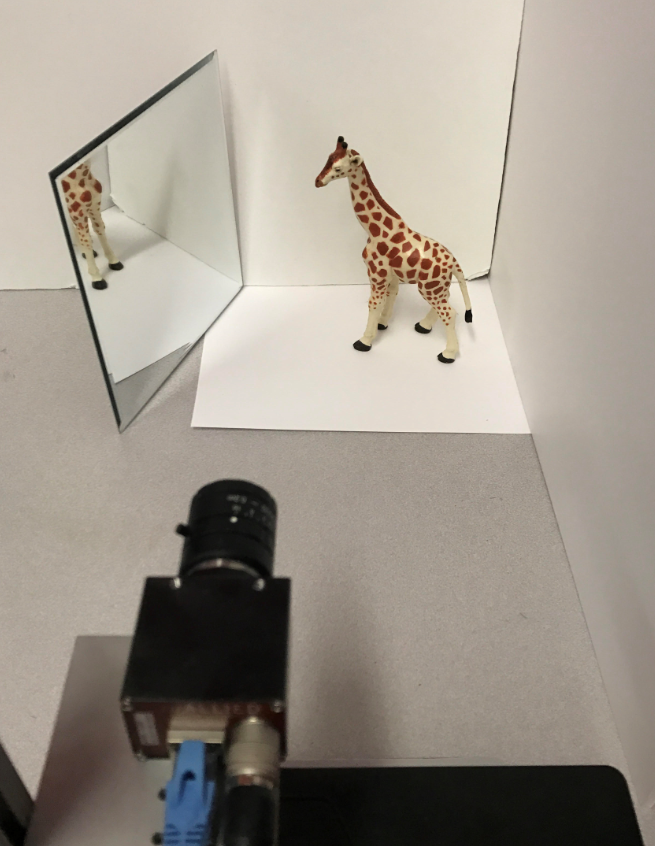}
	\caption{Mirror experiment setup.}
	\label{mirror}
\end{figure}

\begin{figure*}[!t]
	\centering\includegraphics[width=4.30in]{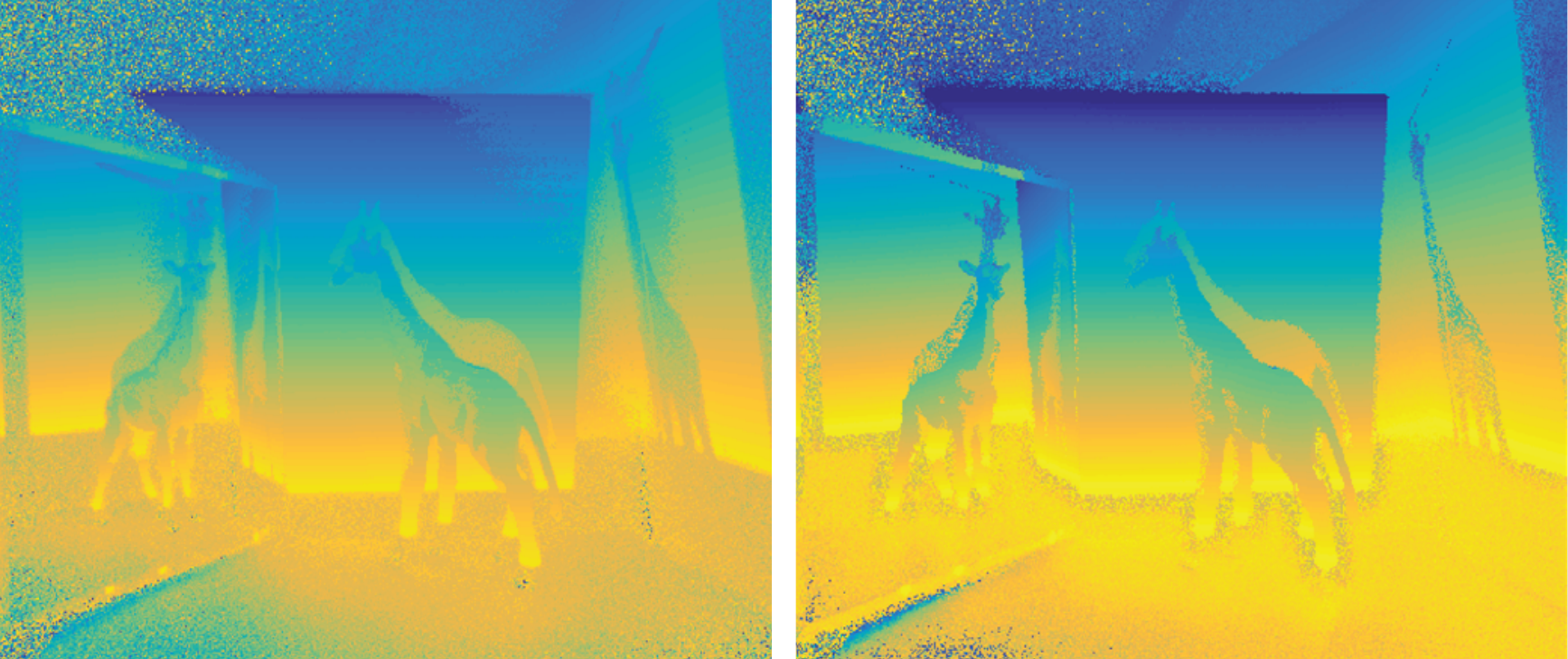}
	\caption{The (left) raw and (right) multi-path phase images of a plastic giraffe.}
	\label{mirrorresu}
\end{figure*}

As a final demonstration, we used a mirror to reflect light from off to on target, a plastic giraffe figurine, as illustrated in the photograph of Fig.~\ref{mirror} with phase results shown in Fig.~\ref{mirrorresu}.  Looking at the raw phase image versus multi-path reconstructed, there are substantial artifacts in the raw phase as indicated by posterization, most visible in the region of the giraffe's neck/chest facing the mirror and especially in the top-right corner of the background screen and on the right side wall. These posterization effects are also visible in the reflected image of the mirror.

\section{Conclusions}
\noindent In this paper, we introduced a novel procedure for extracting the bimodal multipath phase terms for a PMP structured light scan based on modeling the change in magnitude of the observed phasors caused by modulating the spatial frequency of the projected PMP patterns. Furthermore, we introduced the first PMP scans to employ zero-frequency PMP patterns as a way to measure the magnitude in the observed phasors sans multipath. As demonstrated here, the proposed technique is especially geared toward step edges and scanning through semi-transparent surfaces; however, the proposed derivation can be expanded to include more than two paths, although additional investigation is necessary to gauge how practical doing so is. 

Although not considered here, the problem of multi-texture is very similar to the multi-path problem.  Here, a single pixel sees a continuous smooth surface, but the surface texture has a discontinuity or step edge mid-way across the pixel's field of view.  We can define the brighter side of the edge as the foreground surface while the darker side of the edge as the background surface. This means that the phase values inside the foreground surface will have a greater weight, per unit area, than the background surface.  And this has the effect of pushing the combined vector closer to the foreground phase than the background.  While the change may not be as severe as the multi-path problem, the solution is the same, by taking advantage of the presented interesting cue of measuring the constructive and destructive interference between the two light paths. 


{\small
\bibliographystyle{ieee}
\bibliography{egbib}
}

\end{document}